\pdfoutput=1
\documentclass[11pt]{article}

\usepackage[preprint]{acl}

\usepackage{times}
\usepackage{latexsym}

\usepackage[T1]{fontenc}
\usepackage[utf8]{inputenc}

\usepackage{microtype}

\usepackage{inconsolata}

\usepackage{graphicx}

\usepackage{booktabs}
\usepackage{multirow}
\usepackage{array}
\usepackage{xspace}
\usepackage{enumitem}
\usepackage{tabularx}
\usepackage{colortbl}

\usepackage{soul}
\usepackage{bbding}
\newcommand{\datasetname}{\textsc{CharToM-QA}\xspace}
\title{\emph{The Essence of Contextual Understanding in Theory of Mind}:\\ A Study on Question Answering with Story Characters}

\author{\textbf{Chulun Zhou\textsuperscript{1}\thanks{Equal contribution.\quad\quad $\dagger$: Co-corresponding authors.}}~~~
    \textbf{Qiujing Wang\textsuperscript{2}$^{*}$~~~
    \textbf{Mo Yu\textsuperscript{2}$^{\dagger}$}}~~~\textbf{Xiaoqian Yue\textsuperscript{1}}~~~\textbf{Rui Lu\textsuperscript{1}}\\
    \textbf{Jiangnan Li\textsuperscript{2}},~~~\textbf{Yifan Zhou\textsuperscript{2}}~~~\textbf{Shunchi Zhang\textsuperscript{2}}~~~\textbf{Jie Zhou\textsuperscript{2}}~~~\textbf{Wai Lam\textsuperscript{1}$^{\dagger}$}\\
    \textsuperscript{1} The Chinese University of Hong Kong\\
    \textsuperscript{2}Pattern Recognition Center, WeChat AI, Tencent Inc, China \\
    \texttt{\{clzhou,wlam\}@se.cuhk.edu.hk},~~~\texttt{\{yuexiaoqian,luruihfbp\}@link.cuhk.edu.hk}~~~\\~~~\texttt{qw26@njit.edu},~~~\texttt{szhan256@jhu.edu},\\~~~\texttt{\{moyumyu,jiangnanli,geniuszhou,withtomzhou\}@tencent.com}  \\
  }
\begin{document}
\maketitle

\begin{abstract}
Theory-of-Mind (ToM) is a fundamental psychological capability that allows humans to understand and interpret the mental states of others. Humans infer others’ thoughts by integrating causal cues and indirect clues from broad contextual information, often derived from past interactions. In other words, human ToM heavily relies on the understanding about the backgrounds and life stories of others. Unfortunately, this aspect is largely overlooked in existing benchmarks for evaluating machines' ToM capabilities, due to their usage of short narratives without global context, especially personal background of characters. In this paper, we verify \emph{the importance of comprehensive contextual understanding about personal backgrounds in ToM} and assess \emph{the performance of LLMs in such complex scenarios}. To achieve this, we introduce {\datasetname} benchmark, comprising 1,035 ToM questions based on characters from classic novels.
Our human study reveals a significant disparity in performance: the same group of educated participants performs dramatically better when they have read the novels compared to when they have not. In parallel, our experiments on state-of-the-art LLMs, including the very recent o1 and DeepSeek-R1 models, show that LLMs still perform notably worse than humans, despite that they have seen these stories during pre-training. This highlights the limitations of current LLMs in capturing the nuanced contextual information required for ToM reasoning.
\footnote[1]{Our data can be downloaded at Github \url{https://github.com/Encyclomen/CharToM-QA} and Huggingface \url{https://huggingface.co/datasets/ZeroXeno/CharToM-QA}.}
\end{abstract} 

\begin{figure}[t]
  \includegraphics[width=\columnwidth]{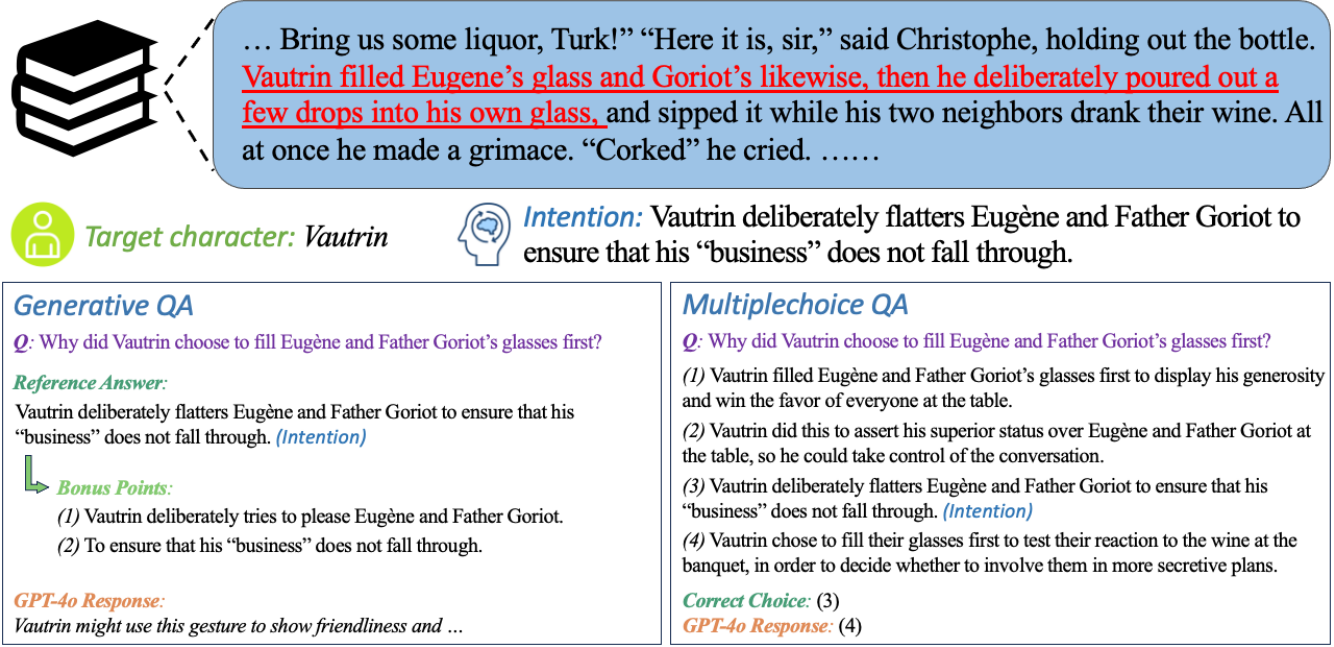}
  \caption{An example from \datasetname benchmark. LLMs make their responses given a question and corresponding story plot in generative and multichoice QA settings.}
  \label{fig:intro}
  %\vspace{-15pt}
\end{figure}

\section{Introduction}
Theory of Mind (ToM) is a psychological term that refers to the process to understand oneself or others by ascribing mental states (typically including the dimensions of belief, intention, emotion, desire, etc) to them \cite{premack1978does}. For human, ToM holds an important position in daily social interactions, as it enables people to explain and predict the behavior of others with no direct access to their minds~\cite{apperly2010mindreaders}.

Recently, a range of emerging Large Language Models (LLMs) have shown remarkable performance in solving complex tasks and generating human-like language. 
To further enhance LLMs' capability to understand human, it is fascinating to explore the extent to which LLMs capture the ToM capabilities and ultimately to equip them with human-level ToM.
To this end, researchers have developed numerous benchmarks to evaluate the ToM capabilities of LLMs~\cite{DBLP:conf/cogsci/GrantNG17,DBLP:conf/emnlp/NematzadehBGGG18,DBLP:conf/emnlp/SapRCBC19,DBLP:conf/emnlp/LeBN19,DBLP:conf/nips/GandhiFGG23,DBLP:conf/emnlp/WuHJM0D23,DBLP:conf/emnlp/0002SZ0K0S23,DBLP:journals/corr/abs-2302-02083,DBLP:journals/corr/abs-2310-03051,DBLP:conf/eacl/ShapiraLAZCGSS24,DBLP:conf/acl/XuZZD024,DBLP:conf/emnlp/ChanJYDF0L0WS24,DBLP:conf/acl/Wu0DSMH24,DBLP:conf/acl/0002WZWBJCHLXH24}.
These benchmarks simulate the experimental design in psychological or cognitive research~\cite{wimmer1983beliefs,dyck2001autism}, and contain elaborately crafted, story-based tasks that target ToM dimensions originally defined in psychology.

Despite their wide coverage of ToM dimensions, these existing benchmarks all largely overlook the importance of \textbf{comprehensive contextual understanding} in human ToM.
In daily lives, humans understand the minds of others based on long historical contexts of their personal backgrounds, rather than with only local circumstances on the spot. This is natural as human behaviors and mental states are also greatly influenced by global background factors of a person, such as social relationships, interactions with others, their personality and past experiences. However, the testing scenarios in existing ToM datasets are usually short of capturing these factors, as they are typically depicted with brief text pieces consisting of only superficial character actions and surrounding environment.

In this paper, we propose \textbf{\datasetname} benchmark (``\textsc{Char}'' \emph{stands for} ``Character'') that evaluates the capability of LLMs on understanding ToM of characters in famous novel books. In \datasetname, the task takes the form of ToM-related question answering (QA) about characters within story plots. This setting naturally addresses the aforementioned challenges of most existing datasets due to the intrinsic features of story plots in novel books: \textbf{1)} diverse social scenarios; \textbf{2)} rich in complex social relationships and interactions; \textbf{3)} high relevance to the whole book storyline. Thus, it alleviates heavy reliance on pre-determined rules to generate testing scenarios and raises higher requirements for comprehensively understanding context when evaluating ToM capability of current LLMs.

For the dataset creation, we adopt an AI-assisted human annotation strategy that exploits publicly available book notes written by online reading app users. As users read, they can underline a specific text fragment and write their comments as book notes. These notes are often closely related to the selected fragments and their surrounding plots, containing valuable interpretation, analysis and thoughts of readers. We first collect a large number of user notes and their linked text from novel books. Then, as in most previous ToM assessments~\cite{DBLP:conf/emnlp/NematzadehBGGG18,DBLP:conf/emnlp/SapRCBC19,DBLP:conf/lrec/TraceyRC0DDDGMM22,DBLP:conf/nips/GandhiFGG23,ma-etal-2023-towards-holistic,DBLP:conf/emnlp/WuHJM0D23}, we keep the notes reflecting certain ToM dimensions of appointed characters, including belief, intention, emotion and desire. Based on these notes, GPT-4o is leveraged to generate ToM descriptions of the four dimensions about given characters. Afterwards, these descriptions are used to construct ToM-related questions with answers. Finally, the experiments are conducted in both generative and multichoice QA settings, where models are asked to make their responses given a question and corresponding story plot, as shown in Figure~\ref{fig:intro}. During the whole process, strict validation procedures are carried out by experts with strong literature background.

For evaluation, in generative QA, the qualities of responses are assessed based on the annotated answers. Particularly, traditional token-based metrics (\emph{e.g.} Rouge~\cite{lin2004rouge}) and embedding-based metrics (\emph{e.g.} cosine similarity between sentence embeddings) cannot give a fine-grained assessment that reflects how well a response matches the answer. Therefore, we design an evaluation protocol that mimics the process of grading papers. Specifically, we extract critical points from the answers, which are expected to be included in responses as bonus points. Then, GPT-4o is used as evaluator for assessment based on these bonus points. Besides, the evaluator also serves like a criticizer to point out defects existing in responses as penalty if there is any inappropriate statement. The bonus point coverage and penalty rate are used to comprehensively assess the qualities of model responses. In multichoice QA, we elaborately construct distraction choices and measure the model accuracy.

We experiment with several current popular LLMs in both generative and multichoice QA settings, where different lengths of plot window are exposed to models. The experimental results show that GPT-4o consistently outperforms other LLMs across the four ToM dimensions in both settings. Moreover, human studies are conducted using multichoice QA. The results reveal that (1) Humans who have read the book greatly outperform the state-of-the-art LLM, while those who have not perform only on par with it. (2) Human ToM largely benefits from familiarity with novels. (3) Humans achieve higher accuracy after reading longer contexts. In contrast, the performances of LLMs almost stay stable with different lengths of plot window. This indicates that the ToM capability of current LLMs still struggles at dealing with complex scenarios and capturing nuanced historical contextual information for robust ToM comprehension.
%We experiment with several popular LLMs, including GPT-4o, GPT-3.5-Turbo-1106, Llama-3.1-8B-Instruct~\cite{dubey2024llama}, Qwen2-7B-Instruct~\cite{yang2024qwen2}, Mistral-7B-Instruct-v0.3~\cite{DBLP:journals/corr/abs-2310-06825}, and InternLM2-7B-Chat~\cite{cai2024internlm2}.

\section{Related Work}
\label{sec:related}
%\paragraph{ToM Benchmarks.}
With the advent of LLMs, many benchmarks have been developed to evaluate the ToM capability of LLMs by simulating the cognitive experiments originally in psychology~\cite{DBLP:conf/cogsci/GrantNG17,DBLP:conf/emnlp/NematzadehBGGG18,DBLP:conf/emnlp/SapRCBC19,DBLP:conf/emnlp/LeBN19,DBLP:conf/nips/GandhiFGG23,DBLP:journals/corr/abs-2302-02083,DBLP:conf/emnlp/WuHJM0D23,DBLP:conf/emnlp/0002SZ0K0S23,DBLP:conf/nips/GandhiFGG23,DBLP:journals/corr/abs-2310-03051,tominamc,DBLP:conf/eacl/ShapiraLAZCGSS24,DBLP:conf/acl/XuZZD024,DBLP:conf/emnlp/ChanJYDF0L0WS24,DBLP:conf/acl/Wu0DSMH24,DBLP:conf/acl/0002WZWBJCHLXH24}. \citet{DBLP:conf/emnlp/NematzadehBGGG18} proposed \textsc{ToMi} for evaluating the capability of models to reason about beliefs. Based on \textsc{ToMi}, \citet{DBLP:conf/emnlp/SapRCBC19} introduced \textsc{Social IQA} that tests social and emotional intelligence of models. \citet{DBLP:conf/emnlp/WuHJM0D23} constructed \textsc{Hi-ToM} to assess higher-order ToM capability of models that involves recursive reasoning on others’ beliefs. \citet{DBLP:conf/emnlp/0002SZ0K0S23} build \textsc{FANTOM} to stress-test ToM within conversational contexts. \citet{DBLP:conf/acl/XuZZD024} constructed \textsc{OpenToM} for assessing ToM with narrative stories containing explicit personality traits or preferences.
\citet{DBLP:conf/acl/Wu0DSMH24} created \textsc{COKE} consisting of cognitive chains that evaluate machine ToM capability to comprehend human activities.
\citet{DBLP:conf/acl/0002WZWBJCHLXH24} introduced \textsc{ToMBench} that encompasses 8 tasks and 31 abilities in social cognition. 
%\textsc{Tom-in-Mac}\cite{tominamc} assessed meta-learning capability of machine ToM by designing the task of few-shot character understanding.
%\citet{DBLP:conf/acl/Wu0DSMH24} created \textsc{COKE} dataset consisting of cognitive chains that evaluate machine ToM capability to comprehend human mental activities and behavioral/affective responses. 

Despite existing ToM benchmarks have covered most dimensions defined in psychology, the constructing procedures and testing scenarios of these datasets are still hindered by inherent limitations. \textbf{1)} The story generation procedures of these benchmarks rely heavily on pre-determined rules and templates~\cite{DBLP:conf/emnlp/NematzadehBGGG18,DBLP:conf/emnlp/LeBN19,DBLP:conf/emnlp/SapRCBC19,DBLP:conf/emnlp/WuHJM0D23}, constraining it from simulating diverse scenarios in reality. This also increases the risk of incorporating predictable regularities and spurious correlations into datasets, bringing about so-called “Clever Hans” phenomenon~\cite{DBLP:journals/corr/abs-1902-10178}. \textbf{2)} The testing scenarios in these datasets are usually short of complex social relationships and interactions so that the importance of comprehensive contextual understanding is almost ignored. Appendix~\ref{sec:appendix_benchmark_comparison} shows a more detailed comparison of \datasetname benchmark with existing benchmarks.

\section{Problem Formulation of ToM in Global Contexts}
\label{sec:problem_formulation}
\paragraph{Problem Formulation}
Similar to existing works on story understanding~\cite{kovcisky2018narrativeqa,pang-etal-2022-quality,xu-etal-2022-fantastic,yu2023personality}, our task adopts a question-answering (QA) format.
We denote the \emph{global context} of a book as $G$, which in practice can be the list of all consecutive paragraphs of the book.
Each QA pair $(q,a)$ is associated with a \emph{plot window} $W \subset G$, which is a book snippet.
The task is then to answer the question according to $W$, with the necessary usage of the contextual knowledge from $G$\footnote[2]{It is worth noting that, like~\cite{kovcisky2018narrativeqa,pang-etal-2022-quality}, the problem allows to infer the thoughts of book characters with plots either after $W$ (\emph{retrospect}), or before $W$ (\emph{predict}).
Both types of contexts naturally occur as part of the long-dependency reasoning in human ToM in daily life.}:
\begin{equation}
%\small
% \begin{aligned}
    P(a\vert q, W, G).\label{eq:problem}
% \end{aligned}
\end{equation}
\paragraph{ToM Dimensions Studied}
In this paper, we consider a set of ToM dimensions prevalently studied in previous work in Section~\ref{sec:related}, \emph{Belief}, \emph{Intention}, \emph{Emotion} and \emph{Desire}. The importance of these dimensions in daily lives have been discussed in~\cite{apperly2010mindreaders}. 
We follow the standard definitions of these dimensions in~\cite{apperly2010mindreaders}, with the detailed definitions provided in Appendix~\ref{sec:appendix_tom_definition_studied}.

\section{\datasetname: Assessing the Contextual Aspect of ToM}
\label{sec:main}
We create the \datasetname benchmark consisting of questions about the minds of appointed characters in novel stories. The benchmark is designed to assess the ToM capability of LLMs in understanding characters within global contexts. The definitions of the four studied ToM dimensions are given in Appendix~\ref{sec:appendix_tom_definition_studied}. The questions and answers are annotated by four human experts in literature, with an AI-assisted annotation strategy that leverages massive publicly available book notes from online reading platforms.

The dataset construction undergoes three steps: (1) book notes collection \& filtering; (2) key note extraction \& paraphrasing; (3) question generation \& verification. Our dataset supports evaluation in both generative and multichoice QA settings.

\begin{figure}[t]
  \includegraphics[width=\columnwidth]{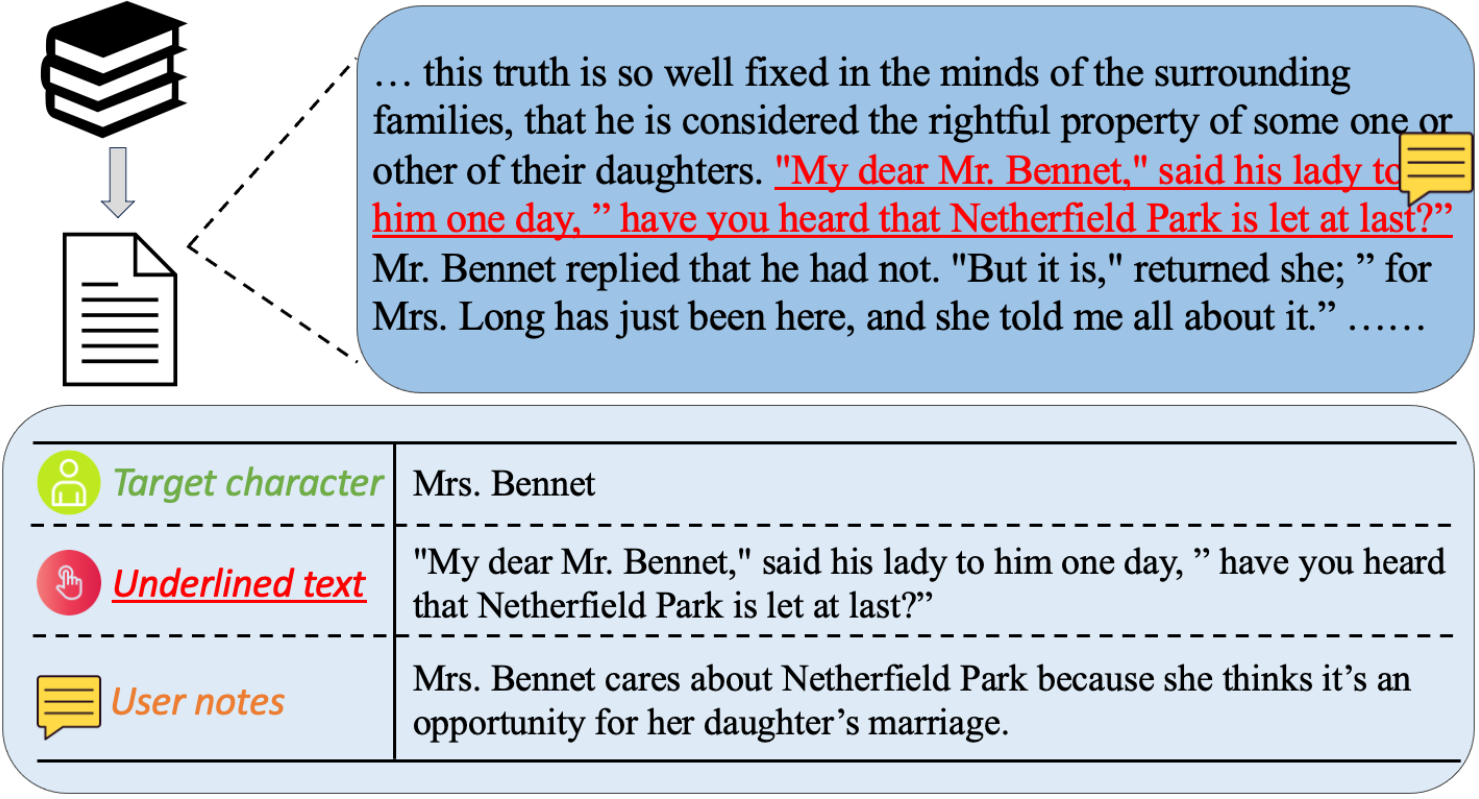}
  \caption{Illustration of notes from users on reading apps. A user underlines a text fragment (\textcolor{red}{red}) and writes his note about the character ``Mrs. Bennet''.}
  \label{fig:user_notes}
\end{figure}
\subsection{Book Notes Collection \& Filtering}
In recent online reading apps (\emph{e.g.} Douban, Kindle), users can underline a text fragment and take notes while reading, as shown in Figure~\ref{fig:user_notes}. These notes usually contain valuable interpretation, analysis and thoughts of readers about the selected fragment and corresponding story plot. 
Following \cite{DBLP:conf/acl/WanMNM19,yu2023personality}, we use the user notes as a proxy to assist the human annotation of our dataset. 
We first download a set of public books available in the Gutenberg project and use their Chinese-translated versions with valid usage licenses. Then, we collect a large number of user notes and their linked text fragment from the Internet with a list of main characters appearing in each book. The books with top 20 most user notes are chosen, as listed in Appendix~\ref{sec:appendix_book_list}.

Next, we filter the notes and keep those that can express any one of the four ToM dimensions of characters within story plots. The prompt used for note filtering is detailed in Appendix~\ref{sec:appendix_prompt_note_filtering}. To ensure the validity of these filtered notes, we also manually check their correctness. Specifically, we sample 100 notes from four books with two annotators who are familiar with the books. For each note, the annotators are required to judge whether it appropriately describes the ToM of the appointed character. The results show that 1) 80\% notes are judged as accurate; 2) 13\% of the notes are open to subjective interpretation by different readers, yet they remain consistent with the book's content and cannot be falsified; 3) only 7\% notes are inappropriate, which demonstrates the overall validity of our filtered notes. Overall, 93\% of notes can be treated as accurate.

\subsection{Construction of Generative QA Task}
\begin{figure}[t]
  \includegraphics[width=\columnwidth]{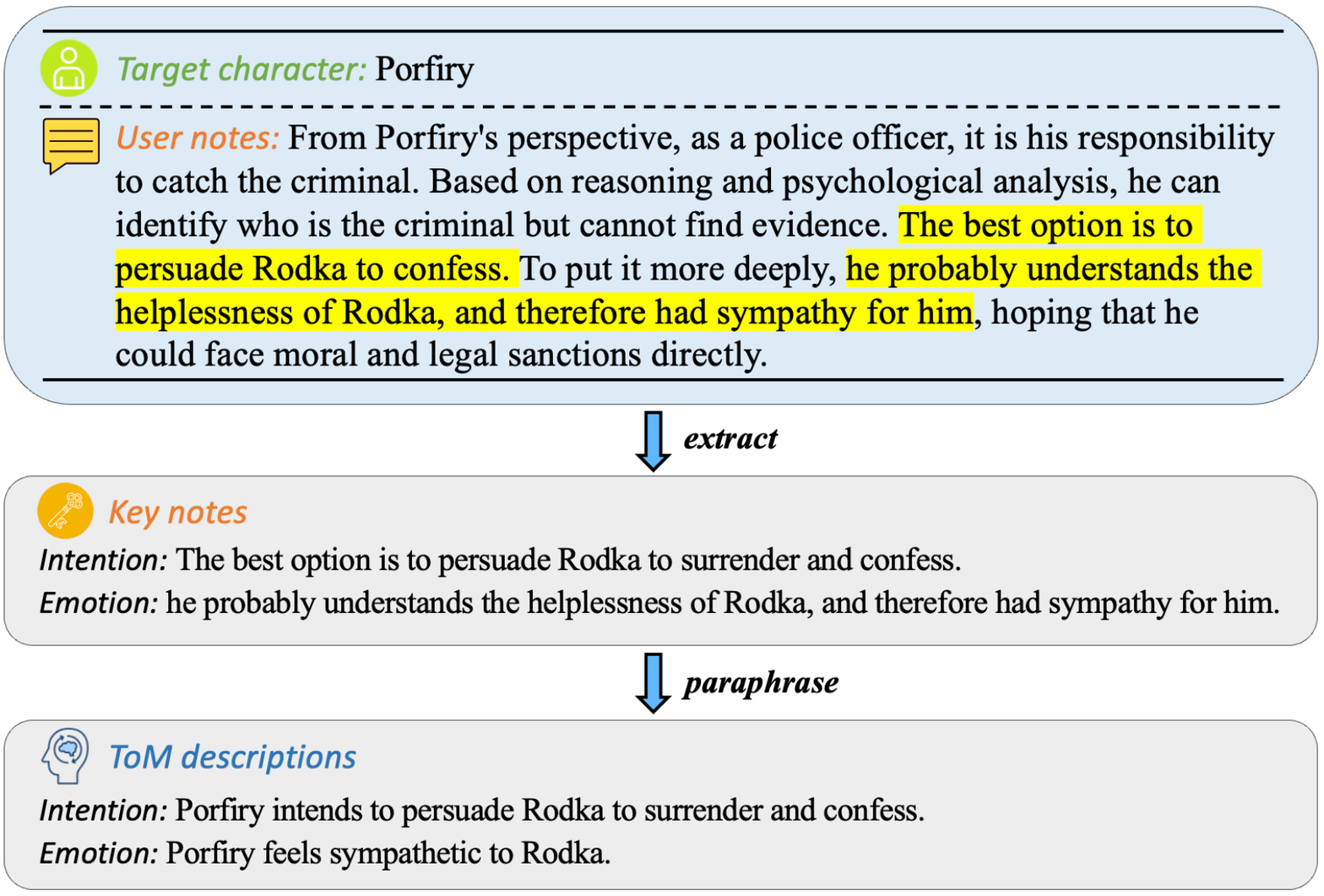}
  \caption{The extract-then-paraphrase approach to fetch ToM descriptions of a target character. The highlighted text pieces are the key notes extracted by annotators, which are then paraphrased into complete statements.}
  \label{fig:extract_paraphrase}
  %\vspace{-10pt}
\end{figure}

\paragraph{Key Note Extraction \& Paraphrasing (\emph{i.e.}, Answer Generation).}
\label{sec:tom_extraction_paraphrasing}
After collection and filtering, we have obtained a pool of highly ToM-related user notes. From these notes, we aim to fetch concrete ToM descriptions of characters within story plots. Specifically, we take a two-stage approach consisting of extraction and paraphrasing operations, as the case in Figure~\ref{fig:extract_paraphrase}.

At the extraction stage, the annotators are instructed to extract the critical part of a note that explicitly reflects one of the four ToM dimensions of a target character, which we call as ``\emph{key note}''. The basic principle of this stage is that annotators should only consider the literal meaning of a note with no additional inference. We elaborately craft examples for every ToM dimension and write a detailed manual to guide annotators. An annotator is only considered qualified after passing a labeling trial. They are given necessary information linked with a user note, including the target character, the underlined text and its surrounding context. After extraction, the key notes are kept for each ToM dimension. Nevertheless, the extracted key notes are often in a form of fragmented expressions that cannot be immediately used as complete ToM descriptions. Thus, a paraphrasing stage is required to convert fragmented parts into complete statements. For each note, we prompt GPT-4o to conduct paraphrasing based on key notes. The details of our used prompt template are presented in Appendix~\ref{sec:appendix_prompt_paraphrasing}. Annotators also check the paraphrased descriptions and make minor edit if necessary.

Although it is feasible for us to merely ask annotators or GPT-4o to directly produce ToM descriptions using the user notes, the advantages of such extract-then-paraphrase operation lie in the following two aspects. First, it relieves the burden of annotators to write complete descriptions totally from scratch. Second, GPT-4o can be more focused on the critical parts of notes without being distracted by other trivial details, reducing the possibility to produce unwanted content.

\begin{figure*}[t]
\setlength{\abovecaptionskip}{6pt}
\setlength{\belowcaptionskip}{-5pt}
\centering
\includegraphics[width=1.0 \textwidth]{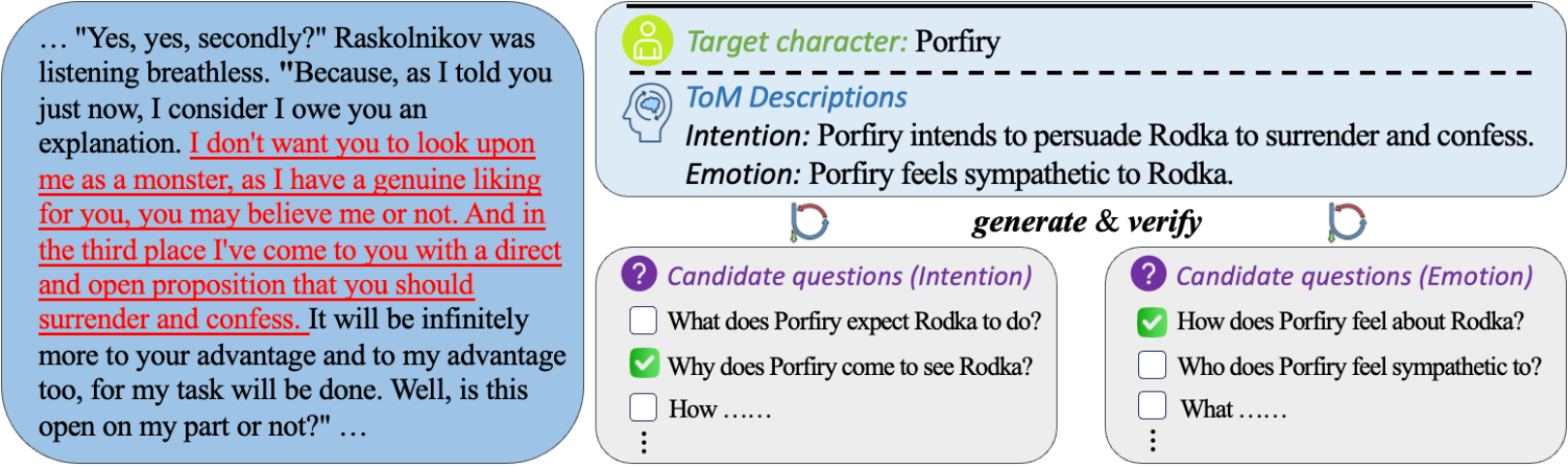}
\caption{Given the corresponding story plot, GPT-4o is used to generate candidate questions about the target character using ToM descriptions. For each dimension, at most 4 candidates are kept after verification. Then, annotators manually choose the best one from these candidates.
}
%\vspace{-5pt}
\label{fig:question_generation}
\end{figure*}

\paragraph{Question Generation \& Verification.}
\label{sec:question_generation_verification}
In this step, we aim to construct ToM-related questions using those ToM descriptions. Figure~\ref{fig:question_generation} depicts the process of question generation. GPT-4o serves as a question generator to produce a set of candidate questions asking about target characters. During generation, along with the ToM description, we also provide GPT-4o with corresponding underlined text and its surrounding context of current story plot. For each ToM description, GPT-4o is asked to generate several candidate questions that adhere to the following requirements:
% \vspace{-13pt}
\begin{itemize}[noitemsep,nolistsep,leftmargin=*]
  \setlength{\itemsep}{0pt}
    \item The ToM description of target character should directly be the fluent and logically correct answer to the question.
    \item The question must be closely related to current story plot.
    \item The question should focus on core content rather than trivial details in the plot.
    %\item The forms of generated questions are diverse.
\end{itemize}
\vspace{0pt}

Subsequently, a verification procedure is involved to preliminarily filter out inappropriate candidates that violate the above requirements.  We find that it is better not to let GPT-4o be aware of the story plot. This is because GPT-4o tends to build trivial connections between question and answer using any possible detail in the story regardless of importance, preserving many low-quality candidates. The prompt templates used for generation and verification are given in Appendix~\ref{sec:appendix_prompt_question_generation}. In this way, we keep at most 4 candidate questions for each ToM description. Then, annotators are responsible for choosing the best one out of these candidates, where slight modifications by human are allowed during examination.

\subsection{Construction of Multichoice QA Task}
Besides the aformentioned Generative QA setting, we support multichoice QA evaluation with elaborately constructed distraction choices for each question.
% In \datasetname, to support multichoice QA evaluation, we elaborately construct distraction choices for each question. 
Given the question-answer pair within a story plot, GPT-4o is used to produce distraction choices that are seemingly plausible but actually unreasonable. Appendix~\ref{sec:appendix_construct_difficult_distraction} provides the details of constructing such difficult distraction choices. Finally, a question is paired with four choices including the correct answer and three incorrect ones, as shown in the Appendix~\ref{sec:appendix_illustrative_cases}. 

\section{Evaluation Protocols for Generative QA}
\label{sec:generative_qa}
Until now, we have acquired ToM-related question-answer pairs along with their linked story plots, which can be used to conduct evaluation in a generative QA setting. Models are given story plots and asked questions to make responses whose qualities are assessed using corresponding answers as reference. Conventionally, token-based metrics (\emph{e.g.} Rouge~\cite{lin2004rouge}) and embedding-based metrics (\emph{e.g.} cosine similarity between sentence embeddings) are mainly used for assessing generative QA tasks. However, these two types of metrics have their inherent limitations. For token-based metrics, they cannot deal with the cases where \textbf{1)} same meanings are expressed with different tokens; \textbf{2)} non-critical tokens dominate in a sentence. Meanwhile, embedding-based metrics only coarsely measure the semantic similarity but cannot explicitly indicate the pros and cons of a response. 

Therefore, inspired by the process of grading papers, we design an evaluation protocol that inspects the bonus points and penalty of a model response. Specifically, for each question, we extract the critical points of its reference answer with the assistance of GPT-4o, which are expected to be included as bonus points in a response. The core requirements for bonus point extraction are: \textbf{1)} Bonus points must be derived from reference answers with no hallucination; \textbf{2)} Different bonus points should orient to different aspects of an answer. The statistics about the bonus points of questions in each dimension is given in Appendix~\ref{sec:appendix_bp_statistics}. During evaluation, a GPT-4o evaluator measures the coverage of bonus points as an indicator of response quality. Moreover, it also criticizes the defects in a response as penalty if there is any inappropriate statement. In this way, both bonus point coverage and penalty rate are used as a comprehensive assessment of response quality. The results in Appendix~\ref{sec:appendix_results_and_human_correlation} also demonstrate that our designed evaluation protocol is more correlated with human judgments than conventional metrics. Particularly, the prompts for extracting bonus points and conducting such evaluation protocol are detailed in Appendices~\ref{sec:appendix_prompt_extract_bonus_points} and~\ref{sec:appendix_prompt_generative_QA_evaluation}, respectively. Appendix~\ref{sec:appendix_illustrative_cases} gives illustrative cases of such evaluation protocol.

%\begin{table}[]
%\begin{center}
%\small
%\begin{tabular}{lcccc} \hline
%\toprule
%\hline
%\vspace{2pt}
%\hline
% & belief & intention & emotion & desire \\ \hline
% \#$Question$ & 201 & 220 & 463 & 151 \\
% \#$Bonus\ Point$ & 311 & 367 & 700 & 212 \\
% \#$Q_{\#bp=1}$ & 108 & 106 & 267 & 95 \\
% \#$Q_{\#bp=2}$ & 77 & 84 & 159 & 51 \\
% \#$Q_{\#bp>2}$ & 16 & 30 & 37 & 5 \\
%\hline
%\end{tabular}
%\vspace{-5pt}
%\caption{Statistics of questions and the distribution of their bonus points in each ToM dimension.
%}
%\label{tab:data_statistics}
%\end{center}
%\vspace{-20pt}
%\end{table}
%\paragraph{Statistics.}
%\datasetname consists of ToM-related questions about characters in the dimensions of belief, intention, emotion and desire. In Table~\ref{tab:data_statistics}, we present the statistics regarding the number of questions and their bonus points in each dimension. Since most questions have less than 3 bonus points, we group them into ``$Q_{\#bp=1}$'', ``$Q_{\#bp=2}$'' and ``$Q_{\#bp>2}$'', which refers to questions with one, two and more than two bonus points.

\section{Experiments}

\begin{table*}[t]
\small
\begin{center}
\begin{tabular}{l|ccc|ccc|ccc|ccc}
%\toprule
\hline
%\noalign{\vskip 3pt}
%\multicolumn{13}{c}{\resizebox{1.2cm}{!}{\textbf{English}}}
\multicolumn{13}{c}{\textbf{Bonus Point Coverage\%} (English) $\uparrow$}
%\vspace{3pt}
\\ \hline
\multicolumn{1}{c|}{\multirow{2}{*}{Model}} & \multicolumn{3}{c|}{Belief}  & \multicolumn{3}{c|}{Intention} & \multicolumn{3}{c|}{Emotion} & \multicolumn{3}{c}{Desire} \\\cline{2-13}
\multicolumn{1}{r|}{} & c=0 & c=1k & c=2k & c=0 & c=1k & c=2k & c=0 & c=1k & c=2k & c=0 & c=1k & c=2k \\ \hline
GPT-4o & \textbf{41.4} & \textbf{43.4} & \textbf{44.4} & \textbf{34.3} & \textbf{35.7} & \textbf{34.6} & \textbf{34.0} & \textbf{37.5} & \textbf{36.1} & \textbf{51.9} & \textbf{50.0} & \textbf{52.8}  \\
GPT-3.5-Turbo-1106 & 36.0 & 38.6 & 35.4 & 29.4 & 30.5 & 31.3 & 28.1 & 31.0 & 32.0 & 47.2 & 43.9 & 42.5  \\
Llama-3.1-8B-Instruct & 30.2 & 37.0 & 37.9 & 24.5 & 25.3 & 30.0 & 28.7 & 30.3 & 31.0 & 35.8 & 38.7 & 40.1 \\
Qwen2-7B-Instruct & 42.1 & 40.2 & 40.8 & 26.4 & 27.8 & 31.1 & 26.0 & 31.4 & 32.7 & 40.1 & 47.2 & 39.6  \\
Mistral-7B-Instruct-v0.3 & 33.4 & 38.9 & 34.1 & 23.2 & 24.3 & 27.5 & 25.4 & 26.4 & 26.4 & 38.7 & 37.3 & 38.7  \\
InternLM2-7B-Chat & 25.4 & 32.8 & 28.6 & 17.7 & 28.1 & 24.0 & 21.9 & 27.4 & 26.3 & 34.4 & 34.9 & 35.4 \\
\hline
%\noalign{\vskip 3pt}
%\multicolumn{13}{c}{\resizebox{1.2cm}{!}{\textbf{English}}}
\multicolumn{13}{c}{\textbf{Penalty Rate\%} (English) $\downarrow$}
%\vspace{3pt}
\\
\hline
\multicolumn{1}{c|}{\multirow{2}{*}{Model}} & \multicolumn{3}{c|}{Belief}  & \multicolumn{3}{c|}{Intention} & \multicolumn{3}{c|}{Emotion} & \multicolumn{3}{c}{Desire} \\\cline{2-13}
\multicolumn{1}{r|}{} & c=0 & c=1k & c=2k & c=0 & c=1k & c=2k & c=0 & c=1k & c=2k & c=0 & c=1k & c=2k \\ \hline
GPT-4o & \textbf{65.7} & \textbf{50.7} & \textbf{51.7} & \textbf{70.0} & \textbf{59.1} & \textbf{60.0} & \textbf{69.1} & \textbf{61.4} & \textbf{61.1} & \textbf{60.3} & \textbf{52.3} & \textbf{47.7}  \\
GPT-3.5-Turbo-1106 & 75.6 & 66.7 & 69.7 & 79.1 & 72.3 & 73.2 & 81.9 & 74.9 & 77.1 & 66.9 & 71.5 & 70.2  \\
Llama-3.1-8B-Instruct & 88.1 & 85.6 & 83.6 & 91.4 & 77.7 & 84.1 & 84.9 & 81.9 & 82.3 & 83.4 & 80.1 & 82.1 \\
Qwen2-7B-Instruct & 83.6 & 79.1 & 80.6 & 85.0 & 83.2 & 85.0 & 89.0 & 84.9 & 80.1 & 84.1 & 78.8 & 79.5  \\
Mistral-7B-Instruct-v0.3 & 81.6 & 72.6 & 75.1 & 85.0 & 83.2 & 80.9 & 89.4 & 84.2 & 83.2 & 80.1 & 83.4 & 75.5  \\
InternLM2-7B-Chat & 86.1 & 84.6 & 86.6 & 94.1 & 82.7 & 84.5 & 92.2 & 86.8 & 85.3 & 90.1 & 84.1 & 86.1
\\ \hline
%\noalign{\vskip 3pt}
%\multicolumn{13}{c}{\resizebox{1.2cm}{!}{\textbf{Chinese}}}
\multicolumn{13}{c}{\textbf{Bonus Point Coverage\%} (Chinese) $\uparrow$}
%\vspace{3pt}
\\ \hline
\multicolumn{1}{c|}{\multirow{2}{*}{Model}} & \multicolumn{3}{c|}{Belief}  & \multicolumn{3}{c|}{Intention} & \multicolumn{3}{c|}{Emotion} & \multicolumn{3}{c}{Desire} \\\cline{2-13}
\multicolumn{1}{r|}{} & c=0 & c=1k & c=2k & c=0 & c=1k & c=2k & c=0 & c=1k & c=2k & c=0 & c=1k & c=2k \\ \hline
GPT-4o & \textbf{50.2} & \textbf{51.4} & \textbf{51.1} & \textbf{38.4} & \textbf{42.5} & \textbf{40.9} & \textbf{42.6} & \textbf{41.6} & \textbf{43.7} & \textbf{59.4} & \textbf{58.5} & \textbf{61.3}  \\
GPT-3.5-Turbo-1106 & 37.3 & 37.3 & 36.7 & 25.9 & 24.3 & 22.6 & 32.6 & 36.7 & 35.4 & 45.3 & 41.5 & 45.3  \\
Llama-3.1-8B-Instruct & 33.4 & 34.4 & 36.0 & 20.7 & 21.3 & 22.9 & 31.7 & 30.4 & 32.1 & 35.8 & 44.3 & 40.1 \\
Qwen2-7B-Instruct & 35.7 & 40.5 & 39.9 & 24.5 & 25.1 & 29.2 & 37.3 & 35.6 & 36.3 & 48.6 & 50.5 & 46.2  \\
Mistral-7B-Instruct-v0.3 & 30.5 & 32.2 & 30.2 & 13.1 & 21.3 & 18.8 & 25.7 & 27.3 & 27.3 & 33.5 & 36.3 & 37.7  \\
InternLM2-7B-Chat & 35.0 & 35.4 & 32.8 & 26.7 & 26.2 & 22.3 & 22.0 & 20.6 & 19.9 & 38.2 & 37.3 & 37.3 
\\ \hline
%\noalign{\vskip 3pt}
%\multicolumn{13}{c}{\resizebox{1.2cm}{!}{\textbf{Chinese}}}
\multicolumn{13}{c}{\textbf{Penalty Rate\%} (Chinese) $\downarrow$}
%\vspace{3pt}
\\
\hline
\multicolumn{1}{c|}{\multirow{2}{*}{Model}} & \multicolumn{3}{c|}{Belief}  & \multicolumn{3}{c|}{Intention} & \multicolumn{3}{c|}{Emotion} & \multicolumn{3}{c}{Desire} \\\cline{2-13}
\multicolumn{1}{r|}{} & c=0 & c=1k & c=2k & c=0 & c=1k & c=2k & c=0 & c=1k & c=2k & c=0 & c=1k & c=2k \\ \hline
GPT-4o & \textbf{48.3} & \textbf{43.3} & \textbf{35.3} & \textbf{58.2} & \textbf{41.4} & \textbf{45.0} & \textbf{45.8} & \textbf{40.6} & \textbf{39.5} & \textbf{43.7} & \textbf{33.8} & \textbf{33.1}  \\
GPT-3.5-Turbo-1106 & 56.2 & 52.7 & 49.3 & 70.0 & 58.6 & 58.6 & 57.9 & 48.8 & 49.7 & 51.0 & 46.4 & 44.4  \\
Llama-3.1-8B-Instruct & 63.7 & 54.2 & 55.7 & 78.2 & 68.2 & 65.9 & 56.6 & 51.4 & 52.2 & 66.2 & 53.0 & 58.3 \\
Qwen2-7B-Instruct & 63.2 & 57.7 & 52.2 & 72.7 & 60.0 & 61.4 & 52.9 & 44.7 & 44.7 & 54.3 & 47.7 & 43.0  \\
Mistral-7B-Instruct-v0.3 & 71.6 & 64.2 & 61.2 & 82.3 & 73.6 & 68.2 & 65.0 & 56.8 & 57.5 & 70.2 & 64.2 & 60.9  \\
InternLM2-7B-Chat & 68.7 & 62.7 & 61.7 & 74.1 & 69.1 & 64.1 & 57.5 & 54.4 & 52.3 & 57.0 & 49.0 & 45.7
\\ \hline
\end{tabular}
\vspace{-5pt}
\caption {Generative QA performances of LLMs in terms of bonus point coverage and penalty rate. For each ToM dimension, the context lengths of story plots exposed to LLMs are set to 0, 1k and 2k tokens.}
\label{tab:performance_generative}
\end{center}
\vspace{-10pt}
\end{table*}
\subsection{Setup}
We conduct experiments on our generative and multichoice QA tasks in English and Chinese. Models are instructed to respond with vanilla prompts as in Appendix~\ref{sec:appendix_prompt_respond_questions}. %We also explore the effect of Chain-of-Thought (CoT) prompting on model performances.
To investigate the capability of the model to exploit contextual information for ToM comprehension, we vary the context lengths of story plots exposed to models. 

For human study, we recruit 8 native Chinese-speaking graduate students to do multichoice questions on a subset of books and use their results for human-LLMs comparison. 
Because it is almost infeasible for a person to write answers to all questions, we only acquire human baselines in multichoice QA on a subset, as in \cite{DBLP:conf/acl/ThaiCKI22}. 

\paragraph{Evaluated LLMs}
In our experiments, we totally evaluate six current popular LLMs, including GPT-4o, GPT-3.5-Turbo-1106, Llama-3.1-8B-Instruct~\cite{dubey2024llama}, Qwen2-7B-Instruct~\cite{yang2024qwen2}, Mistral-7B-Instruct-v0.3~\cite{DBLP:journals/corr/abs-2310-06825}, and InternLM2-7B-Chat~\cite{cai2024internlm2}. We strictly call official APIs or download model weights from Huggingface\footnote[3]{https://huggingface.co/models} repositories with no violation of their terms.

\begin{table*}[t]
\small
\begin{center}
\begin{tabular}{l|ccc|ccc|ccc|ccc}
%\toprule
\hline
%\noalign{\vskip 3pt}
%\multicolumn{13}{c}{\resizebox{1.2cm}{!}{\textbf{English}}}
\multicolumn{13}{c}{\textbf{Accuracy\%} (English)}
%\vspace{3pt}
\\ \hline
\multicolumn{1}{c|}{\multirow{2}{*}{Model}} & \multicolumn{3}{c|}{Belief}  & \multicolumn{3}{c|}{Intention} & \multicolumn{3}{c|}{Emotion} & \multicolumn{3}{c}{Desire} \\\cline{2-13}
\multicolumn{1}{r|}{} & c=0 & c=1k & c=2k & c=0 & c=1k & c=2k & c=0 & c=1k & c=2k & c=0 & c=1k & c=2k \\ \hline
GPT-4o & \textbf{58.7} & \textbf{56.7} & \textbf{54.7} & \textbf{52.3} & \textbf{53.6} & \textbf{51.4} & \textbf{53.1} & \textbf{51.0} & \textbf{51.8} & \textbf{56.3} & \textbf{56.6} & \textbf{55.0}  \\
GPT-3.5-Turbo-1106 & 48.3 & 42.3 & 41.8 & 41.4 & 44.1 & 41.8 & 46.2 & 47.7 & 50.1 & 46.4 & 49.3 & 47.0  \\
Llama-3.1-8B-Instruct & 38.8 & 34.8 & 33.8 & 39.1 & 35.9 & 37.3 & 47.5 & 47.7 & 47.3 & 40.4 & 45.7 & 42.4 \\
Qwen2-7B-Instruct & 44.3 & 41.3 & 36.8 & 38.2 & 37.3 & 36.8 & 46.0 & 46.7 & 47.1 & 44.4 & 39.1 & 41.7  \\
Mistral-7B-Instruct-v0.3 & 38.8 & 36.8 & 38.8 & 36.4 & 35.5 & 33.6 & 41.7 & 43.4 & 45.6 & 41.1 & 43.7 & 43.7  \\
InternLM2-7B-Chat & 39.8 & 34.3 & 35.3 & 30.0 & 36.8 & 35.0 & 41.0 & 43.4 & 45.1 & 45.0 & 44.4 & 40.4 \\
\hline
%\noalign{\vskip 3pt}
%\multicolumn{13}{c}{\resizebox{1.2cm}{!}{\textbf{English}}}
\multicolumn{13}{c}{\textbf{Accuracy\%} (Chinese)}
%\vspace{3pt}
\\
\hline
\multicolumn{1}{c|}{\multirow{2}{*}{Model}} & \multicolumn{3}{c|}{Belief}  & \multicolumn{3}{c|}{Intention} & \multicolumn{3}{c|}{Emotion} & \multicolumn{3}{c}{Desire} \\\cline{2-13}
\multicolumn{1}{r|}{} & c=0 & c=1k & c=2k & c=0 & c=1k & c=2k & c=0 & c=1k & c=2k & c=0 & c=1k & c=2k \\ \hline
GPT-4o & \textbf{52.2} & \textbf{52.2} & \textbf{54.7} & \textbf{50.0} & \textbf{47.3} & \textbf{50.5} & \textbf{51.8} & \textbf{52.1} & \textbf{53.8} & \textbf{54.3} & \textbf{51.7} & \textbf{53.6} \\
GPT-3.5-Turbo-1106 & 37.8 & 37.3 & 40.8 & 33.6 & 37.3 & 41.4 & 43.2 & 47.7 & 49.0 & 35.8 & 41.1 & 34.4  \\
Llama-3.1-8B-Instruct & 37.3 & 35.8 & 39.8 & 37.3 & 38.6 & 35.9 & 43.6 & 45.6 & 44.9 & 37.7 & 35.8 & 40.4 \\
Qwen2-7B-Instruct & 43.3 & 39.3 & 41.1 & 38.2 & 35.5 & 48.7 & 46.0 & 46.0 & 49.0 & 35.8 & 39.7 & 41.1  \\
Mistral-7B-Instruct-v0.3 & 36.8 & 30.8 & 32.3 & 28.6 & 31.8 & 32.7 & 39.1 & 39.7 & 42.1 & 27.2 & 35.1 & 35.8  \\
InternLM2-7B-Chat & 38.3 & 40.3 & 41.8 & 37.3 & 31.4 & 33.2 & 44.7 & 47.3 & 50.5 & 41.1 & 39.7 & 40.4  
\\ \hline
\end{tabular}
\vspace{-5pt}
\caption {Multichoice QA performances of LLMs in terms of accuracy with vanilla prompting. For each ToM dimension, the context lengths of story plots exposed to LLMs are set to 0, 1k and 2k tokens.}
\label{tab:performance_multichoice}
\end{center}
\vspace{-10pt}
\end{table*}

\paragraph{Metrics}
For generative QA, as discussed in Section~\ref{sec:generative_qa}, we exploit GPT-4o as evaluator to assess the qualities of model responses using our proposed evaluation protocol. In our protocol, Bonus Point Coverage (BPC) and Penalty Rate (PR) assess response quality from different perspectives. BPC is calculated as the percentage of bonus points that are judged as being included in responses. PR is the rate of model responses that are judged as defective. Note that a response that includes all bonus points can still contain defects. During evaluation, as a necessary constraint, LLMs are required to produce responses of roughly the same length as corresponding answers. Otherwise, in extreme cases, LLMs could produce arbitrarily long responses to cover as many bonus points as possible. For multichoice QA, we just use the standard accuracy.

%Therefore, response-answer length ratio (RALR), computed as average ratio between the number of tokens in responses and that in corresponding answers, is also an important metric. 

\subsection{Main Results with LLMs}
In this section, we show the performance of LLMs on generative QA and multichoice QA of our \datasetname benchmark.
\paragraph{Generative QA.}
For generative QA, LLMs are evaluated with vanilla prompting in English and Chinese. Table~\ref{tab:performance_generative} gives the results in terms of BPC and PR. For fair comparison, LLMs are required to produce responses whose token numbers are no more than 1.5 times longer than those of answers. It can be seen that GPT-4o achieves the best performances across all dimensions in both English and Chinese. But it only covers roughly 35-50\% (at most 58.5\%) bonus points in every dimension. Meanwhile, PRs of all models are mostly at very high levels, reflecting the non-negligible existence of defects in model responses. After inspection, we find that some models tend to produce excessively long responses that are truncated due to the length constraint. Besides, producing longer responses would increase the risk of including inappropriate statements, which also exacerbates the phenomenon of high PR. Moreover, longer context exposed to models mostly reduce PRs of these models but fail to bring consistent improvements in term of BPC, which will be further discussed in Section~\ref{subsec:analysis}. These results indicate the difficulty of our benchmark and there is still large room of improvement for current LLMs. 
%In Appendix, we also report the response-answer length ratios of these models without truncation in Table~\ref{tab:length_ratio}.

%Qwen2-7B-Instruct and GPT-4o mostly achieve highest BPCs in English and Chinese experiments, respectively. But they only cover roughly 40-50\% (at most 58.5\%) bonus points in every dimension. Meanwhile, PRs of all models are mostly at very high levels, reflecting the non-negligible existence of defects in model responses. The response-answer length ratios of these models are also reported in Table~\ref{tab:length_ratio}. In English experiments, we can observe that the RALR of Qwen2-7B-Instruct is considerably larger than other models. Generally, generating longer responses would increase the risk of including inappropriate statements. To some extent, it brings about the outcome that Qwen2-7B-Instruct covering the most bonus points underperforms GPT-4o in term of PR. In contrary, GPT-4o achieves the best or at least competitive performances across English and Chinese experiments while remaining a relatively stable RALR. The impact of different context lengths exposed to models will be elaborately discussed in Section~\ref{subsec:analysis}. These results indicate the difficulty of our benchmark and there is still large room of improvement for all models.
%\vspace{-8pt}
\paragraph{Multichoice QA.}
For multichoice QA, Table~\ref{tab:performance_multichoice} gives the performances of all models in term of accuracy. It can be seen that GPT-4o also outperforms other models in both English and Chinese experiments. Similarly, the impact of context lengths is still inconsistent. These outcomes basically echo the results in generative QA experiments. %We can also see that model performances on English multichoice QA are mostly better than those on Chinese. 

\subsection{Human Performance}
\label{subsec:human_performance}
In this section, we conduct human study to explore how humans perform on our task, and to assess the role of long-context dependency in human ToM when understanding fictional characters.
Our study addresses the following research questions:
\begin{itemize}[noitemsep,nolistsep,leftmargin=*]
    \item \emph{RQ1: What is the performance gap between humans with and without knowledge on the historical contexts of the characters?}
    This question explores the key hypothesis of this paper regarding the long-dependency nature of human ToM. To investigate this, we divided the human annotators into two groups: those who have read the books and are familiar with the appointed characters (\textbf{\textit{w/.} history}) and those who have not read the books (\textbf{\textit{w/o.} history}). The performance gap between the two groups highlights the importance of long-context dependency in human ToM.
    \item \emph{RQ2: Can humans outperform LLMs?} We compare the performance of the two human groups described earlier with GPT-4o on the same samples, to understand how human ToM can help to better solve our task.
    \item \emph{RQ3: Can human performance benefit from longer input windows?} To investigate this, each annotator is initially asked to answer a question using a limited plot window (\textbf{c=0}). Then, they are provided with extended plot window and decide whether to revise their choices (\textbf{c=2k}). 
\end{itemize}

To enable direct comparison with LLMs, we use the multichoice QA setting and recruit 8 educated participants, all of whom are either PhD candidates or hold PhD degrees across various disciplines.
We sampled 150 questions from 5 books, ensuring that each question is assigned to one person who had read the book and one who had not. 
Each participant is given an \emph{equal number of questions about books they had read and had not}. Combined with the two variations of plot windows, the participants totally make choices on 600 questions.

The result of this study is presented in Table~\ref{tab:human_performance}, from which we draw the following conclusions:
\begin{itemize}[noitemsep,nolistsep,leftmargin=*]
    \item \emph{Answer to RQ1:} Human ToM is largely impacted by familiarity with the target characters. A gap of $\sim$21\% is observed between the two groups.
    \item \emph{Answer to RQ2:} Humans who had read the book greatly outperform the state-of-the-art LLM, while those who had not perform only on par with it. It suggests that the key advantage of human ToM in our task lies in the ability to integrate the characters' global historial contexts.
    \item \emph{Answer to RQ3:} Humans achieves better accuracy when provided with longer contexts. This conforms to the general expectation that longer contexts can 1) help people familiar with the book recall the exact position of the context within the book; 2) provide additional information for those unfamiliar with the book. In contrast, the LLM does not exhibit a similar pattern.
\end{itemize}

\begin{table}[t]
\begin{center}
\small
\setlength{\tabcolsep}{3pt}
\scalebox{0.8}{
\begin{tabular}{c|cc|cc|cc|cc|cc} 
%\toprule
\hline
\multicolumn{11}{c}{\textbf{Accuracy\%} (Chinese)} \\
\hline
 &\multicolumn{2}{c|}{\textit{w/.} history} & \multicolumn{2}{c|}{\textit{w/o.} history} & \multicolumn{2}{c|}{GPT-4o}  & \multicolumn{2}{c|}{o1} & \multicolumn{2}{c}{DS-R1} \\
 &c=0 & c=2k & c=0 & c=2k & c=0 & c=2k & c=0 & c=2k & c=0 & c=2k \\ 
 %\midrule
 \hline
 all &
\cellcolor[HTML]{5AA2C0} \color{black} 59.3 &
\cellcolor[HTML]{4695C0} \color{black} \textbf{62.0} &
\cellcolor[HTML]{D6E6F4} \color{black} 41.3 &
\cellcolor[HTML]{AFD1E7} \color{black} 48.7 &
\cellcolor[HTML]{B8D5EA} \color{black} 47.3 &
\cellcolor[HTML]{D4E4F4} \color{black} 42.0 &
\cellcolor[HTML]{9CC9E1} \color{black} 51.3 &
\cellcolor[HTML]{CEE0F2} \color{black} 43.3 &
\cellcolor[HTML]{71AFD0} \color{black} 54.0 &
\cellcolor[HTML]{AFD1E7} \color{black} 48.7 \\
\hline
indir. &
\cellcolor[HTML]{7DB8DA} \color{black} 54.8 &
\cellcolor[HTML]{5AA2C5} \color{black} 60.6 &
\cellcolor[HTML]{F7FBFF} \color{black} 33.7 &
\cellcolor[HTML]{DAE8F6} \color{black} 40.4 &
\cellcolor[HTML]{C7DBEF} \color{black} 45.1 &
\cellcolor[HTML]{E7F0FA} \color{black} 37.5 &
\cellcolor[HTML]{BFD8ED} \color{black} 46.2 &
\cellcolor[HTML]{DFEBF7} \color{black} 39.4 &
\cellcolor[HTML]{BFD8ED} \color{black} 50.4 &
\cellcolor[HTML]{DFEBF7} \color{black} 40.2 \\
\hline
%\bottomrule
\end{tabular}
}
\caption{Accuracy of multichoice QA (Chinese) by different participants familiar or unfamiliar with selected books. We compare with GPT-4o, o1 and DS-R1 on the same subset. \emph{indir.} refers to indirect questions.}
\label{tab:human_performance}
\end{center}
\vspace{-10pt}
\end{table}

%\section{Analysis and Discussions}
\subsection{Analysis and Discussions}
\label{subsec:analysis}
\paragraph{Can Chain-of-Thought (CoT) help?}
We further explore whether CoT could bring improvement of ToM comprehension.
For this study, we use the recently released strong OpenAI o1 and DeepSeek-R1 (DS-R1)~\cite{guo2025DeepSeek} models on the human study subset of 150 instances.
As shown in Table~\ref{tab:human_performance}, both o1 and DS-R1 outperform GPT-4o, showing the benefit from stronger reasoning ability. However, they still notably lag behind humans. As the CoT of DS-R1 is visible, we closely examine its generated thoughts on randomly selected 100 samples. We find that 76\% of the thoughts initially recalled either the character's personal background or prior plot events that are relevant to the question. Among the remaining 24\% of cases, three-fourths of questions are about emotions, which is as expected because emotions are more likely to be temporary mental states that rely less on global contexts. These findings further highlights the importance of comprehensive contextual understanding for ToM reasoning, a promising direction for enhancing future LLM.

\paragraph{LLMs Struggles at Exploiting Longer Plot Windows.}
While many studies have shown the superior capability of LLMs in handling long inputs, even up to 100k tokens~\cite{dubey2024llama,hurst2024gpt}, we find they fail to utilize longer plot windows in our tasks.
In both generative and multichoice QA settings, the performances of LLMs almost remain stable as the plot window is enlarged.\footnote[4]{It is worth noting that PRs steadily drop as LLMs read longer plot windows, showing that they do incorporate  contextual information from story plots.}

This failure can be attributed to two potential reasons. First, despite the ability to process long inputs, LLMs struggle with capturing dependencies between paragraphs and enhancing the understanding of individual paragraphs based on these dependencies~\cite{xu-etal-2024-fine}.
Second, the LLMs may solve our task largely with their memory of the books and reviews, exhibiting the \emph{Clever Hans} phenomenon~\cite{DBLP:conf/eacl/ShapiraLAZCGSS24} -- as the evaluated books and their reviews have been exposed to the LLMs during pre-training. It is likely that the LLMs can recall the associated plot details and knowledge with both short and long windows.
We leave this study of these reasons in future work.
% \textcolor{red}{When it comes to the impact of different context lengths exposed to models, their performances do not exhibit consistent changes. In generative QA experiments, as the context length increases, BPCs of these LLMs almost remain stable. This suggests current LLMs still struggle at effectively exploiting contextual information to comprehend ToM of characters in scenarios with complex social relationships and interactions. However, it is worth noting that PRs steadily drop as LLMs read longer context, which shows that these LLMs indeed make use of contextual information in story plots and reduce defects in their responses.} 

\paragraph{LLMs Perform Poorly on Indirect Questions.}
To gain further insight on the LLM performance, we categorize our questions into two types, \emph{direct} and \emph{indirect} questions, for analysis. 
Direct question refers to one with the answer or its synonymous expressions explicitly appear in the given story plot.
We manually labeled the 150 instances used in human study, resulting in 104 indirect questions, with the performance on this subset shown in Table~\ref{tab:human_performance}.
% Direct questions refer to those questions whose direct clues of answers, including the answer itself or synonymous expressions, explicitly appear in the given story plot. Accordingly, the other questions belong to indirect questions. 
%We now investigate how well these LLMs perform on questions that have no direct clue within the given story plot.

As expected, both humans unfamiliar with the books and the LLMs drop dramatically on the subset of indirect questions. Furthermore, it is noteworthy that both o1 and DS-R1 models also perform worse on the indirect questions than on the full set. As indirect questions typically require broader global context of the whole stories rather than only superficial textual clues in the given plot windows, it suggests that the ToM capabilities of current LLMs are still limited especially in the aspect of comprehensive contextual understanding.

\section{Conclusion}
We introduce \datasetname benchmark that aims to evaluate machines' ToM capability in scenarios with complex social relationships and interactions. \datasetname focuses on the aspect of contextual understanding in comprehending ToM of characters in novels, which is largely ignored in existing benchmarks. The tasks take the forms of generative and multichoice QA. For generative QA, we design an evaluation protocol inspecting bonus points and penalty in model responses. Experimental results reveal that current advanced LLMs, including o1 and DS-R1, still lag behind humans and struggle at effectively exploiting context to comprehend ToM of characters. Overall, our \datasetname benchmark highlights the importance of comprehensive contextual understanding for advancing ToM reasoning in LLMs.

\section*{Limitations}
Our proposed \datasetname benchmark is built from the notes written by readers when reading classic novels. For the creation, as we adopt an AI-assisted annotation strategy, the annotation results will be affected by the personal understanding from the annotators and GPT-4o. Thus, it requires the annotators to basically know the story of these books. Besides, the benchmark includes the ToM dimensions of belief, intention, emotion and desire, which are prevalently studied in previous work. However, there are still other ToM dimensions that can be investigated, such as knowledge and thoughts. 

We also design an evaluation protocol that assesses model responses by BPC and PR, where GPT-4o is used as an evaluator. Such evaluation would cost much if the dataset is very large or the stories are excessively long. For those who cannot afford massively calling OpenAI APIs, other locally-deployed open-source LLMs can also be used to make assessments.
\section*{Ethical Consideration}
The construction of \datasetname benchmark involves the usage of public books and notes written by reading app users. During the process of using these data, we strictly adhere to the guidelines in Internet Research Ethics. All the books are available in the Gutenberg project and their Chinese-translated version are properly used with valid license. For the notes, we collect and use them with necessary procedures to protect the privacy of online reading app users who wrote the notes. For annotators participating in the benchmark construction, they are adequately paid according to their working hours. In our human study, we do not collect any personally identifiable information of the participants. The participants are only asked to respond to multichoice questions and not required to provide any new written input. 

\section*{Potential Risks}
As our benchmark is derived from classic novels, the stories are essentially fictional and might have limitations of the time. Meanwhile, the notes written by users reflect their personal interpretation of the story, which may contain content expressing negative emotions. However, please note that the construction approaches and evaluation protocols in this paper can also be used in other type of tasks.
% \section*{Acknowledgments}

% Bibliography entries for the entire Anthology, followed by custom entries
%\bibliography{anthology,custom}
% Custom bibliography entries only
\bibliography{custom}

\appendix
\newpage
\clearpage
\section*{Appendix}

\section{Book List}
\label{sec:appendix_book_list}
We keep the books with top 20 user notes for the construction of \datasetname benchmark. These books are publicly accessible and span different historical periods, genres, and topics. Table~\ref{tab:book_statistics} lists the chosen books and the number of questions belonging to each book in the descending order. It is also notable that our dataset has included a number of unique target characters (150 in total) from these books. These characters are significantly diverse, crafted by different authors with varying literary styles. Particularly, these characters possess distinct personalities, encompassing various types and fundamentally different personal backgrounds. These elements could introduce diversity to our dataset, effectively mitigating potential biases related to literary style and topic.
\begin{table}[ht]
\begin{center}
\small
\begin{tabular}{lc} 
\hline
%\toprule
%\vspace{2pt}
%\hline
 Book & $\# Question$ \\ \hline
 Madame Bovary & 167 \\
 The Count of Monte-Cristo & 101 \\
 Crime and Punishment & 94 \\
 Of Human Bondage & 88 \\
 Pride and Prejudice & 82 \\
 Anna Karenina  & 79 \\
 War and Peace & 53 \\
 Jane Eyre & 49 \\
 Wuthering Heights & 42 \\
 The Brothers Karamazov & 37 \\
 Anne Of Green Gables & 33 \\
 Little Women & 32 \\
 The Idiot & 30 \\
 Twenty Thousand Leagues under the Sea & 29 \\
 Les Miserables & 23 \\
 Notre-Dame de Paris & 22 \\
 Oliver Twist & 21 \\
 Father Goriot & 19 \\
 Tess of the d'Urbervilles & 19 \\
 The Red and the Black & 15 \\
\hline
 Total & 1,035 \\
 \hline
\end{tabular}
\caption{The book list and the number of questions from each book.
}
\label{tab:book_statistics}
\end{center}
\vspace{-10pt}
\end{table}

\section{ToM Dimensions Studied}
\label{sec:appendix_tom_definition_studied}
In this paper, we concern four ToM dimensions of characters within story plots including \emph{Belief}, \emph{Intention}, \emph{Emotion} and \emph{Desire}. The definitions of these dimensions are as follows
\begin{itemize}
  \setlength{\itemsep}{-2.5pt}
    \item \textbf{Belief}: Beliefs encompass both objective facts and subjective perceptions concerning the existence or truth of something.
    \item \textbf{Emotion}: Emotions are strong feelings deriving from one's circumstances, mood, or relationships with others. And emotions are variously associated with thoughts, feelings, behavioral responses, and a degree of pleasure or displeasure.
    \item \textbf{Intention}: Intentions are blueprints that steer actions, encompassing both future plans and the motivations driving current behaviour.
    \item \textbf{Desire}: Desires encompass both physical needs and psychological yearnings. Desires incline people toward action and fulfilling desires is pleasurable. Their fulfillment is normally experienced as pleasurable in contrast to the negative experience of failing to do so. 
\end{itemize}
\vspace{0pt}

\begin{table*}[t]
\small
\begin{center}

\resizebox{1.0\textwidth}{!}{
\begin{tabular}{l|c|ccc|ccc|p{5.8cm}|c}
%\toprule
\hline
\multicolumn{10}{c}{\textbf{Benchmarks \textit{w/.} Template-based Problem Construction}} \\
\hline
 %& \multirow{2}{*}{\textbf{Task Form}} & \multicolumn{3}{c|}{\textbf{Story Creation}} & \multicolumn{3}{c|}{\textbf{Question}\,\textit{\&}\,\textbf{Answer Construction}} & \multirow{2}{*}{\textbf{Character Background}} & \multirow{2}{*}{\textbf{\textit{\#Eval\_Q}}} \\
 & \multirow{2}{*}{Task Form} & \multicolumn{3}{c|}{Story Creation} & \multicolumn{3}{c|}{Problem Construction} & \multirow{2}{*}{Character Background} & \multirow{2}{*}{\textit{\#Eval\_Q}} \\
 \cline{3-8}
 & & Template & AI & Human & Template & AI & Human & & \\
 \hline
 \textsc{ToMi}~\cite{DBLP:conf/emnlp/NematzadehBGGG18} & \emph{MQA} & \Checkmark & & & \Checkmark & & & \emph{N/A} & 6,000 \\
 \textsc{Social IQA}~\cite{DBLP:conf/emnlp/SapRCBC19} & \emph{MQA} & -- & -- & -- & \Checkmark & & \Checkmark & \emph{N/A} & 2,224 \\
 \textsc{Hi-ToM}~\cite{DBLP:conf/emnlp/WuHJM0D23} & \emph{MQA} & \Checkmark & & & \Checkmark & & & \emph{N/A} & 1,000 \\
 \textsc{FanToM}\cite{DBLP:conf/emnlp/0002SZ0K0S23} & \emph{MQA},\emph{GQA} & \Checkmark & \Checkmark & & \Checkmark & \Checkmark & & \emph{N/A} & 2,118 \\
 \textsc{BigToM}~\cite{DBLP:conf/nips/GandhiFGG23} & \emph{MQA} & \Checkmark & \Checkmark &  & \Checkmark & \Checkmark &  & \emph{N/A}  & 5,000 \\
 \textsc{NegotiationToM}~\cite{DBLP:conf/emnlp/ChanJYDF0L0WS24} & \emph{MQA} &  &  & \Checkmark & \Checkmark &  & \Checkmark & \emph{pre-defined character's preference items} & 13,800 \\
 \textsc{OpenToM}~\cite{DBLP:conf/acl/XuZZD024} & \emph{MQA} & \Checkmark & \Checkmark & & \Checkmark & & \Checkmark & \emph{pre-defined personlity traits and preferences} & 16,008 \\ 
 \textsc{COKE}$^{*}$~\cite{DBLP:conf/acl/Wu0DSMH24} & \emph{NLG} & -- & -- & -- & \Checkmark &  \Checkmark & \Checkmark  & \emph{N/A} & - \\
 \hline
 \multicolumn{10}{c}{\textbf{Benchmarks \textit{w/o.} Template-based Problem Construction}} \\
\hline 
& \multirow{2}{*}{Task Form} & \multicolumn{3}{c|}{Story Creation} & \multicolumn{3}{c|}{Problem Construction} & \multirow{2}{*}{Character Background} & \multirow{2}{*}{\textit{\#Eval\_Q}} \\
\cline{3-8}
 & & Template & AI & Human & Template & AI & Human & & \\
 \hline
 %\textsc{COKE}~\cite{DBLP:conf/acl/Wu0DSMH24} & \emph{NLG} & & \Checkmark & \Checkmark &  &  \Checkmark & \Checkmark  & \emph{pre-configured event immediately preceding character's current situation} & - \\
 \textsc{FauxPas-EAI}~\cite{DBLP:conf/acl/ShapiraZG23} & \emph{MQA},\emph{GQA} & &  &\Checkmark &  & & \Checkmark & \emph{N/A} & 80 \\
 \textsc{ToMChallenges}~\cite{DBLP:conf/conll/MaGX23} & \emph{MQA},\emph{GQA} & \Checkmark &  & \Checkmark &  &  & \Checkmark & \emph{N/A} & 360 \\
 \textsc{ToMBench}~\cite{DBLP:conf/acl/0002WZWBJCHLXH24} & \emph{MQA} &  & \Checkmark & \Checkmark &  & \Checkmark & \Checkmark & \emph{N/A} & 2,860 \\
 %\textsc{G4C}~\cite{DBLP:conf/acl/ZhouZHP0C0A23} & \emph{NLG} & & \Checkmark & \Checkmark & & & \Checkmark & \XSolidBrush & \\
 % \textsc{Conan}~\cite{DBLP:conf/acl/ZhaoZX0Z0024} & \emph{Infer} &  & \Checkmark & \Checkmark &  &  & \Checkmark & \emph{character's relationship graphs} & - \\
 \hline
 \datasetname~(Ours) & \emph{MQA},\emph{GQA} & & & \Checkmark & & \Checkmark & \Checkmark & \emph{all of the character's previous experience depicted in story} & 1,035 \\
 %\datasetname~(Ours) & \emph{MQA},\emph{GQA} & & & \Checkmark & & \Checkmark & \Checkmark & \emph{previous experiences and onsite situations depicted in story} & 1,035 \\
\hline
\end{tabular}
}
\vspace{-5pt}
\caption{Comparison among \datasetname and other ToM benchmarks in terms of task form, story generation, problem construction, character background and the number of evaluation questions ($\#Eval\_Q$). In the column of ``Task Form'', \emph{MQA}, \emph{GQA} and \emph{NLG} refer to multichoice QA, generative QA and natural language generation tasks, respectively. ``\Checkmark'' denotes that a benchmark possesses corresponding properties. ``\emph{N/A}'' means the story merely depicts character's current situation without extra character background. ``-'' in the $\#Eval\_Q$ column means the task form does not support the counting of questions equivalent to standard QA settings.
*The COKE and \textsc{Social IQA} datasets are essentially about thoughts on input events/situations so the concept of \emph{stories} does not apply.
}
\label{tab:dataset_comparison}
\end{center}
\vspace{-10pt}
\end{table*}

\section{Comparison of Benchmark Characteristics}
\label{sec:appendix_benchmark_comparison}
In Table~\ref{tab:dataset_comparison}, we compare the characteristics of our \datasetname and other ToM benchmarks. First, our \datasetname provides both multichoice QA and generative QA tasks. Second, as the classic novel stories are directly used, \datasetname needs no additional story creation procedure, unlike most of other benchmarks. Meanwhile, it naturally possesses high-quality diverse social scenarios with complex social relationships and interactions due to the intrinsic features of human-written novels. Third, for the problem construction, most other benchmarks involve templates for generating questions or answers. Although this kind of template-based approach can produce a relatively large number of problems, it could bring shortcuts and spurious correlations into benchmarks. In contrast, we adopt an AI-assisted human annotation strategy elaborated in Section~\ref{sec:main} without any pre-defined template. This simultaneously ensures the efficiency and quality of our annotation process.

Last and most importantly, in the aspect of character background,  \datasetname is the only benchmark that incorporates character's previous experience depicted in story. Most of other ToM benchmarks merely provide depictions of character's current situation or some pre-defined background information such as conversation topics, preference items. As demonstrated by the human study described in Section~\ref{subsec:human_performance}, such personal background is necessary for comprehensive contextual understanding about characters to correctly answer ToM-related questions in \datasetname. Specifically, participants with similar educational levels who have read the book (and thus possess knowledge of long personal backgrounds) perform significantly better than those who have not read when given the same contexts. It shows that our \datasetname highlights the importance of comprehensive contextual understanding for ToM reasoning, an aspect overlooked by previous benchmarks. 

\begin{table}[t]
\begin{center}
\resizebox{.45\textwidth}{!}{
\begin{tabular}{lcccc}
%\toprule
\hline
 & belief & intention & emotion & desire \\ \hline
 \#$Question$ & 201 & 220 & 463 & 151 \\
 \#$Bonus\ Point$ & 311 & 367 & 700 & 212 \\
 \#$Q_{\#bp=1}$ & 108 & 106 & 267 & 95 \\
 \#$Q_{\#bp=2}$ & 77 & 84 & 159 & 51 \\
 \#$Q_{\#bp>2}$ & 16 & 30 & 37 & 5 \\
\hline
\end{tabular}
}
\vspace{-5pt}
\caption{Statistics of questions and their bonus points in each ToM dimension.
}
\label{tab:data_statistics}
\end{center}
\vspace{-10pt}
\end{table}

\section{Statistics of Questions \& Bonus Points}
\label{sec:appendix_bp_statistics}
\datasetname consists of ToM-related questions about characters in the dimensions of belief, intention, emotion and desire. For the evaluation of generative QA, the bonus points for each question are extracted for subsequent assessment. In Table~\ref{tab:data_statistics}, we present the statistics of questions and bonus points in each dimension. Since most questions have less than 3 bonus points, we group them into ``$Q_{\#bp=1}$'', ``$Q_{\#bp=2}$'' and ``$Q_{\#bp>2}$'', which refers to questions with one, two and more than two bonus points.

%\section{Human Correlation of Automatic Evaluation Metrics on Generative QA}
\section{Correlations of Evaluation Metrics with Humans on Generative QA}
\label{sec:appendix_results_and_human_correlation}
For Generative QA, besides the designed LLM-based evaluation protocol, we also use conventional token-based (RougeL) and embedding-based (BertScore) metrics for assessing the quality of model responses. In Table~\ref{tab:results_and_human_correlation}, we report the experimental results in terms of BPC, RougeL and BertScore when the context length is 2k. The reported values are the weighted averages of corresponding metrics over the four ToM dimensions.
\begin{table}[t]
\begin{center}
\resizebox{.48\textwidth}{!}{
\begin{tabular}{lccc}
%\toprule
\hline
 & BPC(\%) & RougeL(\%) & BertScore(\%) \\ \hline
 GPT-4o & \textbf{39.8} & \textbf{21.5} & \textbf{56.6} \\
 GPT-3.5-Turbo-1106 & 34.0 & 19.6 & 51.6 \\
 Llama-3.1-8B-Instruct & 33.5 & 16.2 & 50.8 \\
 Qwen2-7B-Instruct & 34.9 & 15.9 & 50.5 \\
 Mistral-7B-Instruct-v0.3 & 29.9 & 19.0 & 49.7 \\
 InternLM2-7B-Chat & 27.6 & 17.7 & 49.7 \\
\hline
 \emph{Human Correlation*} & \emph{0.92} & \emph{0.72} & \emph{0.81} \\
\hline
\end{tabular}
}
\vspace{-5pt}
\caption{
Generative QA performances of LLMs in terms of BPC, RougeL and BertScore with the context length being 2k. ``*'': The human correlation of these metrics are calculated on the subset used in our human study.
}
\label{tab:results_and_human_correlation}
\end{center}
\vspace{-10pt}
\end{table}
From the table, our LLM-based metric gives similar relative rankings among different LLMs as conventional metrics. Moreover, on the subset used in our human study, we let human experts assess model responses by giving scores ranging from 0 (nonsense) to 1.0 (perfect). The Pearson Correlations of human judgments with BPC, RougeL and BertScore are 0.92, 0.72 and 0.81, respectively. These results further justify the reliability of our designed evaluation protocol.

\section{Prompt for Book Note Filtering}
\label{sec:appendix_prompt_note_filtering}
For book notes filtering, we incorporate the above definitions into the prompt and keep those notes that can reflect any one of the four ToM dimension of characters. The prompt for filtering book notes is presented in Table~\ref{tab:prompt_note_filtering}.
\begin{table}[t]
  \fbox{\begin{minipage}{0.462\textwidth}
  Assuming you are an expert in psychology and literary. Reading app users will take notes while reading novels, which may include the analysis about characters. Please use theory of mind to determine whether the user notes about the novel plots explicitly describe the \$\{\emph{dimension}\} of the target characters in the plot. \\\\
  {[Definition of \$\{\emph{dimension}\}]}: \$\{\emph{definition}\}\\
  {[User Note]}: \$\{\emph{user\_note}\}\\
  {[Target Character]}: \$\{\emph{target\_character}\}\\\\
  If ``Yes'', describe what the \$\{\emph{dimension}\} of \$\{\emph{target\_character}\} is; If not, give your reason. Note that (1) We need the \$\{\emph{dimension}\} of the characters in the novel expressed by readers, not that of the reader. (2) Only consider the content about the \$\{\emph{dimension}\} explicitly described in the note. Do not guess or make any unnecessary prediction.\\\\
  Please answer in the following format:\\
  ``Yes / No''\\
  ``\$\{\emph{dimension}\} of \$\{\emph{target\_character}\} / Reason''
  \end{minipage}}
  \vspace{-5pt}
  \caption{The prompt for book note filtering.}
  \vspace{-10pt}
  \label{tab:prompt_note_filtering}
\end{table}

\section{Prompt for ToM Paraphrasing}
\label{sec:appendix_prompt_paraphrasing}
In Section~\ref{sec:tom_extraction_paraphrasing}, after the key note is extracted, we use GPT-4o to paraphrase it into a complete description about the ToM of the target character within the story plot. The prompt for paraphrasing is presented in Table~\ref{tab:prompt_key_note_paraphrasing}.
\begin{table}[t]
  \fbox{\begin{minipage}{0.462\textwidth}
  Assuming you are an expert in psychology and literary. Reading app users will take notes while reading novels, which may include the analysis about characters. The critical fragments of a user note that can reflect the \$\{\emph{dimension}\} of the target character in the novel plot are extracted as key note. Please use the theory of mind to paraphrase the fragmented key note into a complete ToM description about the \$\{\emph{dimension}\} of \$\{\emph{target\_character}\}. You are also given the original note and its corresponding underlined text.\\\\
  {[Underlined Text]}: \$\{\emph{underlined\_text}\}\\
  {[User Note]}: \$\{\emph{user\_note}\}\\
  {[Key Note]}: \$\{\emph{key\_note}\}\\
  {[Target Character]}: \$\{\emph{target\_character}\}\\\\
   Note that (1) We need the \$\{\emph{dimension}\} of the target character in the novel, not that of the reader. (2) Only consider the content about the \$\{\emph{dimension}\} explicitly described in the note. Do not guess or make any unnecessary prediction. (3) Do not always copy the content of the key note. You are encouraged to make necessary polishing without changing the original meaning. (4) Straightforwardly describe \$\{\emph{dimension}\} and avoid providing additional explanations.\\\\
  The ToM description about the \$\{\emph{dimension}\} of \$\{\emph{target\_character}\} is:
  \end{minipage}}
  \vspace{-5pt}
  \caption{The prompt for key note paraphrasing.}
  \label{tab:prompt_key_note_paraphrasing}
  \vspace{-10pt}
\end{table}

\section{Prompt for Question Generation \& Verification}
\label{sec:appendix_prompt_question_generation}
In Section~\ref{sec:question_generation_verification}, we leverage GPT-4o to generate a set of candidate questions, to which the ToM description of the target character is the answer. Then, at most 4 candidate questions are kept for each ToM description after verification. The prompts for question generation and verification are presented in Table~\ref{tab:prompt_question_generation} and Table~\ref{tab:prompt_question_verification}, respectively.

\begin{table*}[t]
  \begin{tabular}{|p{\textwidth}|}
  \hline
Assuming you are an expert in psychology and literary. Please raise questions related to the \$\{\emph{dimension}\} of the characters in the novel based on [Story Plot] and [Underlined Text] from the book \$\{\emph{book\_name}\}.\\\\
  {[Story Plot]}: \$\{\emph{story\_plot}\}\\
  {[Underlined Text]}: \$\{\emph{underlined\_text}\}\\
  {[Target Character]}: \$\{\emph{target\_character}\}\\
  {[ToM Description]}: \$\{\emph{ToM\_description}\}\\\\
   Note that (1) {[ToM Description]} should directly be the fluent and logically correct answer to your question. (2) The questions should be closely related to current story plot. (3) The question should focus on core content rather than trivial details in the plot. (4) Only raise one question.\\\\
  The question you raise that meets the above requirements is:\\
  \hline
  \end{tabular}
  \vspace{-5pt}
  \caption{The prompt for question generation.}
  \label{tab:prompt_question_generation}
  %\vspace{-10pt}
\end{table*}

\begin{table*}[t]
  \begin{tabular}{|p{\textwidth}|}
  \hline
Assuming you are an expert in psychology and literary. 
  You are reading the book \$\{\emph{book\_name}\}. There are a piece of description about the \$\{\emph{dimension}\} of \$\{\emph{target\_character}\} and a question related to the book story. Please judge whether {[Candidate Question]} is an appropriate question whose direct answer is {[ToM description]}.\\\\
  {[Candidate Question]}: \$\{\emph{candidate\_question}\}\\
  {[ToM description]}: \$\{\emph{ToM\_description}\}\\\\
  An appropriate question should satisfy the following requirements: (1) {[Candidate Question]} should not be too broad. (2) [ToM description] should be a fluent and logically correct answer to {[Candidate Question]}.\\\\
  Please answer in the following format:\\
  ``Appropriate / Inappropriate''\\
  ``Reason''\\
  \hline
  \end{tabular}
  \vspace{-5pt}
  \caption{The prompt for question verification.}
  \label{tab:prompt_question_verification}
  %\vspace{-10pt}
\end{table*}

\section{Prompt for Extracting Bonus Points}
\label{sec:appendix_prompt_extract_bonus_points}
In \datasetname benchmark, for each question, we provide the bonus points that are expected to be included in model responses. The prompt for extracting bonus points is presented in Table~\ref{tab:prompt_bonus_point_extraction}.

\section{Construction of Difficult Distraction Choices for Multichoice QA}
\label{sec:appendix_construct_difficult_distraction}
Furthermore, to increase task difficulty, we also exploit the model responses obtained in the evaluation of generative QA. Concretely, the defects of those responses pointed out by the evaluator are used as potential misleading directions. During the construction of distraction choices, GPT-4o can selectively use these misleading directions for reference to produce more difficult content. In this way, our multichoice QA task becomes much more challenging. Table~\ref{tab:prompt_construct_difficult_distraction} gives the prompt for the construction of such distraction choices.

\section{Prompt for Responding to Questions}
\label{sec:appendix_prompt_respond_questions}
Table~\ref{tab:prompt_respond_questions_gqa} and Table~\ref{tab:prompt_respond_questions_mqa} give the prompts for responding to the questions in generative and multichoice QA during experiments, respectively.
%\vspace{-5pt}

\begin{table*}[t]
  \begin{tabular}{|p{\textwidth}|}
  \hline
  Assuming you are an expert in psychology and literary. Based on your profound understanding of the story in the book \$\{\emph{book\_name}\}, extract the critical bonus points from the {[Answer]} for the {[Question]}. The bonus points are used to comprehensively assess the quality of responses from other respondents. Below are the {[Story Plot]}, {[Question]} about the character \$\{\emph{target\_character}\} and the {[Reference Answer]}. \\\\
  {[Story Plot]}: \$\{\emph{story\_plot}\}\\
  {[Question]}: \$\{\emph{question}\} \\
  {[Reference Answer]}: \$\{\emph{reference\_answer}\}
  \\\\
  Note that (1) Bonus points must be derived from the reference answer without hallucination. Bonus points are the indispensable points in the reference answer, but should not include any unnecessary explanatory content. (2) Different bonus points should orient to different aspects of the reference answer. (3) The semantic granularity of the bonus points should not be too fine. Each bonus point must be a sentence that constitutes a complete semantic unit. Semantically coherent content in the reference answer should not be broken down into multiple bonus points.\\\\
  Please use `{\textbackslash}n' to separate multiple bonus points line by line and every point should conform to the following format:\\
  {[Bonus Point]}: ``Content of bonus point'' \\
  \hline
  \end{tabular}
  \caption{The prompt for extracting bonus points from reference answers.}
  \label{tab:prompt_bonus_point_extraction}
\end{table*}

\begin{table*}[t]
  \begin{tabular}{|p{\textwidth}|}
  \hline
  Assuming you are an expert in psychology and literary. 
  Please construct appropriate multichoice questions for the given [Story Plot] from the the book \$\{\emph{book\_name}\}. \\\\  
  {[Story Plot]}: \$\{\emph{story\_plot}\}\\
  {[Question]}: \$\{\emph{question}\} \\
  {[Reference Answer]}: \$\{\emph{reference\_answer}\} \\
  {[Misleadings]}: \$\{\emph{misleadings}\}\\\\
  Note that (1) Ensure that the reference answer is the only correct answer to the question. (2) The three distraction candidate choices should be seemingly plausible but actually unreasonable. (3) You can refer to the reference answer and misleadings to construct distraction choices. (4) Do not repeat the same content in different distraction choices. \\\\
  Please provide distraction candidate choices that meet the above requirements and conform to the following format:\\
  {[Distraction Choice 1]}: (Choice Content)\\
  {[Distraction Choice 2]}: (Choice Content)\\
  {[Distraction Choice 3]}: (Choice Content) \\
  \hline
  \end{tabular}
  \caption{The prompt for constructing difficult distraction choices in multichoice QA.}
  \label{tab:prompt_construct_difficult_distraction}
\end{table*}

\begin{table*}[t]
  \begin{tabular}{|p{\textwidth}|}
  \hline
  Assuming you are an expert in psychology and literary. Based on your profound understanding of the story in the book \$\{\emph{book\_name}\}, Please answer the [Question].
  The following items give the [Story Plot] and the [Question]. \\\\
  {[Story Plot]}: \$\{\emph{story\_plot}\}\\
  {[Question]}: \$\{\emph{question}\} 
  \\\\
  Note that you should respond to the question using only one concise sentence, with around \$\{\emph{length\_of\_answer}\} tokens. \\
  Your response is:  \\
  \hline
  \end{tabular}
  \caption{The prompt for responding to questions in generative QA.}
  \label{tab:prompt_respond_questions_gqa}
  \vspace{20pt}
\end{table*}

\begin{table*}[t]
  \begin{tabular}{|p{\textwidth}|}
  \hline
  Assuming you are an expert in psychology and literary. Based on your profound understanding of the story in the book \$\{\emph{book\_name}\}, Please answer the [Question].
  The following items give the [Story Plot], the [Question] and the four candidates from [Candidate Choices].. \\\\
  {[Story Plot]}: \$\{\emph{story\_plot}\}\\
  {[Question]}: \$\{\emph{question}\} \\
  {[Candidate Choices]}: \\
  (1). \$\{\emph{cand\_choice\_1}\} \\
  (2). \$\{\emph{cand\_choice\_2}\} \\
  (3). \$\{\emph{cand\_choice\_3}\} \\
  (4). \$\{\emph{cand\_choice\_4}\} \\\\
  Note that \\
  1. You should only choose one candidate from (1),(2),(3),(4). \\
  2. Only output the index of your chosen candidate. \\
  3. Do not include any other unnecessary content or symbol. \\\\
  Your choice is:  \\
  \hline
  \end{tabular}
  \caption{The prompt for responding to questions in multichoice QA.}
  \label{tab:prompt_respond_questions_mqa}
\end{table*}

\section{Prompt for Generative QA Evaluation}
\label{sec:appendix_prompt_generative_QA_evaluation}
For generative QA evaluation, we leverage GPT-4o as an evaluator to assess the quality of model responses based on the bonus points extracted from the reference answer. Table~\ref{tab:prompt_inspect_bonus_point} presents the prompt for inspecting which bonus point is included in a response. Besides, the evaluator also finds defects in a response as penalty. The bonus point coverage and penalty rate are the two metrics measuring the quality of a response. The prompt for finding defects in a response is presented in Table~\ref{tab:prompt_point_out_defect_in_response}.

\begin{table*}[t]
  \begin{tabular}{|p{\textwidth}|}
  \hline
  Assuming you are an expert in psychology and literary. Based on your profound understanding of the story in the book \$\{\emph{book\_name}\}, assess the quality of the response to the given {[Question]} according to the provided {[Reference Answer]} and corresponding {[Bonus Points]}. You are supposed to indicate which bonus points are explicitly or implicitly included in the response.
  \\\\
  {[Question]}: \$\{\emph{question}\}\\
  {[Reference Answer]}: \$\{\emph{reference\_answer}\} \\
  {[Bonus Points]}: \$\{\emph{bonus\_points}\} \\
  {[Response]}: \$\{\emph{response}\}
  \\\\
  Note that (1) The reference answer and bonus points must be the core basis of assessment. (2) The response may be similar to a certain bonus point but with different wording. As long as it expresses a similar meaning to this bonus point with no error, the point can be considered to be validly included. (3) If the response does not include any bonus point, just output `{[Included Bonus Points]}: None'\\\\
  Please conform to the following format: \\
  {[Included Bonus Points]}: ``<Indices of the included bonus points separated by commas, such as `1,2'>'' \\
  \hline
  \end{tabular}
  \caption{The prompt for inspecting which bonus points are included in a response.}
  \label{tab:prompt_inspect_bonus_point}
\end{table*}

\begin{table*}[ht]
  \begin{tabular}{|p{\textwidth}|}
  \hline
  Assuming you are an expert in psychology and literary. Based on your profound understanding of the story in the book \$\{\emph{book\_name}\}, according to the {[Story Plot]} and {[Reference Answer]}, you are supposed to detect whether there is any defect existing in the {[Response]} to the {[Question]}.
  \\\\
  {[Story Plot]}: \$\{\emph{story\_plot}\}\\
  {[Question]}: \$\{\emph{question}\}\\
  {[Reference Answer]}: \$\{\emph{reference\_answer}\} \\
  {[Response]}: \$\{\emph{response}\}
  \\\\
  Note that (1) The reference answer and story plot must be the core basis for detecting defects in the response. (2) Defects refers to factual or logical errors in the response. (3) Do not reluctantly find defects when the response is totally reasonable. If there is no defect, just output `{[Defect]}: None'\\\\
  Please conform to the following format: \\
  {[Defects]}: ``Content of defects in the response''\\
  \hline
  \end{tabular}
  \caption{The prompt for pointing out defects in a response.}
  \label{tab:prompt_point_out_defect_in_response}
\end{table*}

\section{Illustrative Cases}
\label{sec:appendix_illustrative_cases}
Table~\ref{tab:appendix_case_1} gives a case of question about Vautrin's intention from the book ``Father Goriot'', accompanied with the responses of GPT-4o. For the generative QA, the response of GPT-4o covers one of the two bonus points. The penalty explains the defects existing in the response. For the multichoice QA, the model selects the wrong candidate choice. Table~\ref{tab:appendix_case_2} gives another case about Monte Cristo's belief from the book ``The Count of Monte Cristo''. Note that this question involves the second-order belief of the character, which can evaluate the second-order ToM capability of models. We can see that GPT-4o covers both bonus points in generative QA and select the correct choice in multichoice QA.

\begin{table}[t]
    \centering
    \begin{tabular}{|p{0.46\textwidth}|}
    \hline
    \textcolor[HTML]{3C76AF}{\textit{\textbf{Plot Window}}} \\
    \hline
    ...... Bring us some liquor, Turk!'' ``Here it is, sir,'' said Christophe, holding out the bottle. \ul{Vautrin filled Eugene's glass and Goriot's likewise, then he deliberately poured out a few drops into his own glass}, and sipped it while his two neighbors drank their wine. All at once he made a grimace. ``Corked'' he cried. 
    % "The devil!You can drink the rest of this, Christophe, and go and find another bottle; 
    ......  \\ 
    \hline
    \hline
    % \colorbox[HTML]{DDEBFF}{\textbf{Generative QA}} \\
    \textcolor[HTML]{3C76AF}{\textit{\textbf{Generative QA}}}    \\
    \hline
    \textcolor[HTML]{8F6EB0}{\textit{\textbf{Question:}}} Why did Vautrin choose to fill Eugène and Father Goriot's glasses first?   \\
    \hline
    \textcolor[HTML]{469C76}{\textit{\textbf{Reference Answer:}}} Vautrin deliberately flatters Eugène and Father Goriot to ensure that his ``business'' does not fall through.    \\
    % \hline
    \textcolor[HTML]{83CA70}{\textit{\textbf{Bonus Points:}}} \\
    \small
    \textit{\ (1) Vautrin deliberately tries to please Eugène and Father Goriot.} \textcolor[HTML]{83CA70}{\CheckmarkBold} \\
    \small
    \textit{\ (2) To ensure that his ``business'' does not fall through.}\textcolor[HTML]{C76663}{\XSolidBrush}\\
    
    \hline
    \textcolor[HTML]{E08A56}{\textit{\textbf{GPT-4o Response:}}} Vautrin might use this gesture to show friendliness and generosity, thereby fostering closer relationships and easing Eugène's distant attitude toward him, all while concealing his true intentions.\\
    \textcolor[HTML]{C76663}{\textit{\textbf{Penalty:}}} The response mentioning ``easing Eugène's distant attitude toward him'' is neither explicitly stated nor implied in the excerpt. Vautrin's actions are more likely driven by a need to maintain control of the situation and conceal his true intentions, rather than simply addressing Eugène's cold demeanor.\\
    \hline
    \hline
    \textcolor[HTML]{3C76AF}{\textit{\textbf{Multichoice QA}}}   \\
    \hline    
    \textcolor[HTML]{8F6EB0}{\textit{\textbf{Question:}}}Why does Vautrin choose to fill Eugène and Father Goriot's glasses first?    \\
    \textcolor[HTML]{8F6EB0}{\textit{\textbf{Candidate Choices:}}}\\
    \small
    \textit{\ (1)} Vautrin filled Eugène and Father Goriot's glasses first to display his generosity and win the favor of everyone at the table.\\
    \small
    \textit{\ (2)} Vautrin did this to assert his superior status over Eugène and Father Goriot at the table, so he could take control of the conversation.\\
    \small
    \textit{\ (3)} Vautrin deliberately flatters Eugène and Father Goriot to ensure that his ``business'' does not fall through. \\
    \small
    \textit{\ (4)} Vautrin chose to fill their glasses first to test their reaction to the wine at the banquet, in order to decide whether to involve them in more secretive plans.\\
    \hline
    \textcolor[HTML]{469C76}{\textit{\textbf{Correct Choice:}}} (3)  \\
    \hline
    \textcolor[HTML]{E08A56}{\textit{\textbf{GPT-4o Response:}}} (4)\\
    \hline
    \end{tabular}
    \caption{Case 1 from ``Father Goriot''. \textcolor[HTML]{83CA70}{\CheckmarkBold} means the response of GPT-4o covers the bonus point. \textcolor[HTML]{C76663}{\XSolidBrush} means the corresponding bonus point is not covered. The defects of response are also indicated in the penalty. For multichoice QA, the model selects the wrong choice.}
    \label{tab:appendix_case_1}
\end{table}

\begin{table}[t]
    \centering
    \begin{tabular}{|p{0.46\textwidth}|}
    \hline
    \textcolor[HTML]{3C76AF}{\textit{\textbf{Plot Window}}} \\
    \hline
    ...... 
    % Lord Wilmore, having heard the door close after him, returned to his bedroom, where with one hand he pulled off his light hair,
    % his red whiskers, his false jaw, and his wound, to resume the black hair, dark complexion, and pearly teeth of the Count of Monte Cristo.
It was M. de Villefort, and not the prefect, who returned to the house of M. de Villefort. \ul{The procureur felt more at ease, although he had learned nothing really satisfactory, and, for the first time since the dinner-party at Auteuil, he slept soundly.} ......  \\ 
    \hline
    \hline
    % \colorbox[HTML]{DDEBFF}{\textbf{Generative QA}} \\
    \textcolor[HTML]{3C76AF}{\textit{\textbf{Generative QA}}}    \\
    \hline
    \textcolor[HTML]{8F6EB0}{\textit{\textbf{Question:}}} What does the Count of Monte Cristo believe Mr. Villefort's attitude towards him is, and what actions does he expect Villefort to take?   \\
    \hline
    \textcolor[HTML]{469C76}{\textit{\textbf{Reference Answer:}}} The Count of Monte Cristo knows that Mr. Villefort must harbor suspicions about him and will inquire everywhere for information about him.\\
    % \hline
    \textcolor[HTML]{83CA70}{\textit{\textbf{Bonus Points:}}}\\
    \small
    \textit{\ (1) Christophe the Count of Monte Cristo knows that Mr. Villefort must have doubts about him.}\textcolor[HTML]{83CA70}{\CheckmarkBold}\\
    \small
    \textit{\ (2) Mr. Villefort will inquire about news concerning the Count of Monte Cristo everywhere.}\textcolor[HTML]{83CA70}{\CheckmarkBold}\\
    
    \hline
    \textcolor[HTML]{E08A56}{\textit{\textbf{GPT-4o Response:}}} The Count of Monte Cristo believed that Mr. Villefort harbored suspicion towards him and would continue to monitor and investigate his actions through various means.\\
    \textcolor[HTML]{C76663}{\textit{\textbf{Penalty:}}} None\\
    \hline
    \hline
    \textcolor[HTML]{3C76AF}{\textit{\textbf{Multichoice QA}}}   \\
    \hline    
    \textcolor[HTML]{8F6EB0}{\textit{\textbf{Question:}}} What does the Count of Monte Cristo believe Mr. Villefort's attitude towards him is, and what actions does he expect Villefort to take?   \\
    % \hline
    \textcolor[HTML]{8F6EB0}{\textit{\textbf{Candidate Choices:}}}\\
    \small
    \textit{\ (1)} The Count of Monte Cristo believes that Monsieur de Villefort holds a cautious attitude towards him and will take careful actions to observe his movements.\\
    \small
    \textit{\ (2)} The Count of Monte Cristo plans to take a series of actions to seek revenge on Monsieur de Villefort and his family, and therefore believes that Villefort has already realized his plan.\\
    \small
    \textit{\ (3)} The Count of Monte Cristo knows that Mr. Villefort must harbor suspicions about him and will inquire everywhere for information about him.\\
    \small
    \textit{\ (4)} As no specific threat is discovered, the Count of Monte Cristo believes that Monsieur de Villefort will not take any immediate action and will simply maintain a watchful stance for the time being.\\
    
    \hline
    \textcolor[HTML]{469C76}{\textit{\textbf{Correct Choice:}}} (3)   \\
    \hline
    \textcolor[HTML]{E08A56}{\textit{\textbf{GPT-4o Response:}}} (3)\\
    \hline
    \end{tabular}
    \caption{Case 2 from ``The Count of Monte Cristo''. \textcolor[HTML]{83CA70}{\CheckmarkBold} means the response of GPT-4o covers the bonus point. The response of GPT-4o covers all bonus points with no penalty in generative QA. It also selects the correct choice in multichoice QA. Note that this question involves the second-order ToM capability.}
    \label{tab:appendix_case_2}
\end{table}

%\paragraph{Results of Generative QA without truncation.}
%For generative QA, LLMs are evaluated with vanilla prompting in English and Chinese. Table~\ref{tab:performance_generative} gives the results in terms of BPC and PR. It can be seen that Qwen2-7B-Instruct and GPT-4o mostly achieve highest BPCs in English and Chinese experiments, respectively. But they only cover roughly 40-50\% (at most 58.5\%) bonus points in every dimension. Meanwhile, PRs of all models are mostly at very high levels, reflecting the non-negligible existence of defects in model responses. The response-answer length ratios of these models are also reported in Table~\ref{tab:length_ratio}. In English experiments, we can observe that the RALR of Qwen2-7B-Instruct is considerably larger than other models. Generally, generating longer responses would increase the risk of including inappropriate statements. To some extent, it brings about the outcome that Qwen2-7B-Instruct covering the most bonus points underperforms GPT-4o in term of PR. In contrary, GPT-4o achieves the best or at least competitive performances across English and Chinese experiments while remaining a relatively stable RALR. The impact of different context lengths exposed to models will be elaborately discussed in Section~\ref{subsec:analysis}. These results indicate the difficulty of our benchmark and there is still large room of improvement for all models.

\end{document}